\documentclass[lettersize,journal]{IEEEtran}
\usepackage{amsmath,amsfonts}
\usepackage{algorithmic}
\usepackage{algorithm}
\usepackage{array}
\usepackage[caption=false,font=normalsize,labelfont=sf,textfont=sf]{subfig}
\usepackage{textcomp}
\usepackage{stfloats}
\usepackage{url}
\usepackage{verbatim}
\usepackage{graphicx}
\usepackage{cite}
\usepackage[numbers,sort&compress]{natbib}
\usepackage[colorlinks,
            linkcolor=red,
            anchorcolor=blue,
            citecolor=green
            ]{hyperref}
\usepackage{booktabs}
\usepackage{multirow}
\usepackage{multicol}
\usepackage{graphicx}
\usepackage{stfloats}
\usepackage{tikz,xcolor,hyperref}
\definecolor{lime}{HTML}{A6CE39}
\DeclareRobustCommand{\orcidicon}{%
    \begin{tikzpicture}
    \draw[lime, fill=lime] (0,0) 
    circle [radius=0.16] 
    node[white] {{\fontfamily{qag}\selectfont \tiny ID}};    \draw[white, fill=white] (-0.0625,0.095) 
    circle [radius=0.007];    \end{tikzpicture}
    \hspace{-2mm}}
\foreach \x in {A, ..., Z}{%
    \expandafter\xdef\csname orcid\x\endcsname{\noexpand\href{https://orcid.org/\csname orcidauthor\x\endcsname}{\noexpand\orcidicon}}
    }
\begin{document}

\title{BirdNeRF: Fast Neural Reconstruction of Large-Scale Scenes From Aerial Imagery}
\newcommand{\orcidauthorA}{0009-0002-3732-6832}
\newcommand{\orcidauthorB}{0000-0002-6284-1724}
\newcommand{\orcidauthorC}{0000-0002-4443-4367}
\newcommand{\orcidauthorD}{0000-0001-9409-2118}
\author{Huiqing Zhang\orcidA{}, Yifei Xue\orcidC{}, Ming Liao\orcidD{} and Yizhen Lao\orcidB{}*, ~\IEEEmembership{Member,~IEEE}
}

\markboth{IEEE Transactions on Geoscience and Remote Sensing,~Vol.~X, No.~X, August~2024}%
{Shell \MakeLowercase{\textit{et al.}}: A Sample Article Using IEEEtran.cls for IEEE Journals}

\IEEEpubid{0000--0000/00\$00.00~\copyright~2021 IEEE}

\maketitle

\begin{abstract}
In this study, we introduce BirdNeRF, an adaptation of Neural Radiance Fields (NeRF) designed specifically for reconstructing large-scale scenes using aerial imagery. Unlike previous research focused on small-scale and object-centric NeRF reconstruction, our approach addresses multiple challenges, including (1) Addressing the issue of slow training and rendering associated with large models.  (2) Meeting the computational demands necessitated by modeling a substantial number of images, requiring extensive resources such as high-performance GPUs. (3) Overcoming significant artifacts and low visual fidelity commonly observed in large-scale reconstruction tasks due to limited model capacity. Specifically, we present a novel bird-view pose-based spatial decomposition algorithm that decomposes a large aerial image set into multiple small sets with appropriately sized overlaps, allowing us to train individual NeRFs of sub-scene.  This decomposition approach not only decouples rendering time from the scene size but also enables rendering to scale seamlessly to arbitrarily large environments. Moreover, it allows for per-block updates of the environment, enhancing the flexibility and adaptability of the reconstruction process. Additionally, we propose a projection-guided novel view re-rendering strategy, which aids in effectively utilizing the independently trained sub-scenes to generate superior rendering results. 
We evaluate our approach on existing datasets as well as against our own drone footage, improving reconstruction speed by 10x over classical photogrammetry software and 50x over state-of-the-art large-scale NeRF solution, on a single GPU with similar rendering quality.
\end{abstract}

\begin{IEEEkeywords}
NeRF, large-scale reconstruction, aerial image, spatial decomposition, projection-guided.
\end{IEEEkeywords}

\section{Introduction}

\begin{figure}[!t]
    \centering
    \includegraphics[width=1\linewidth]{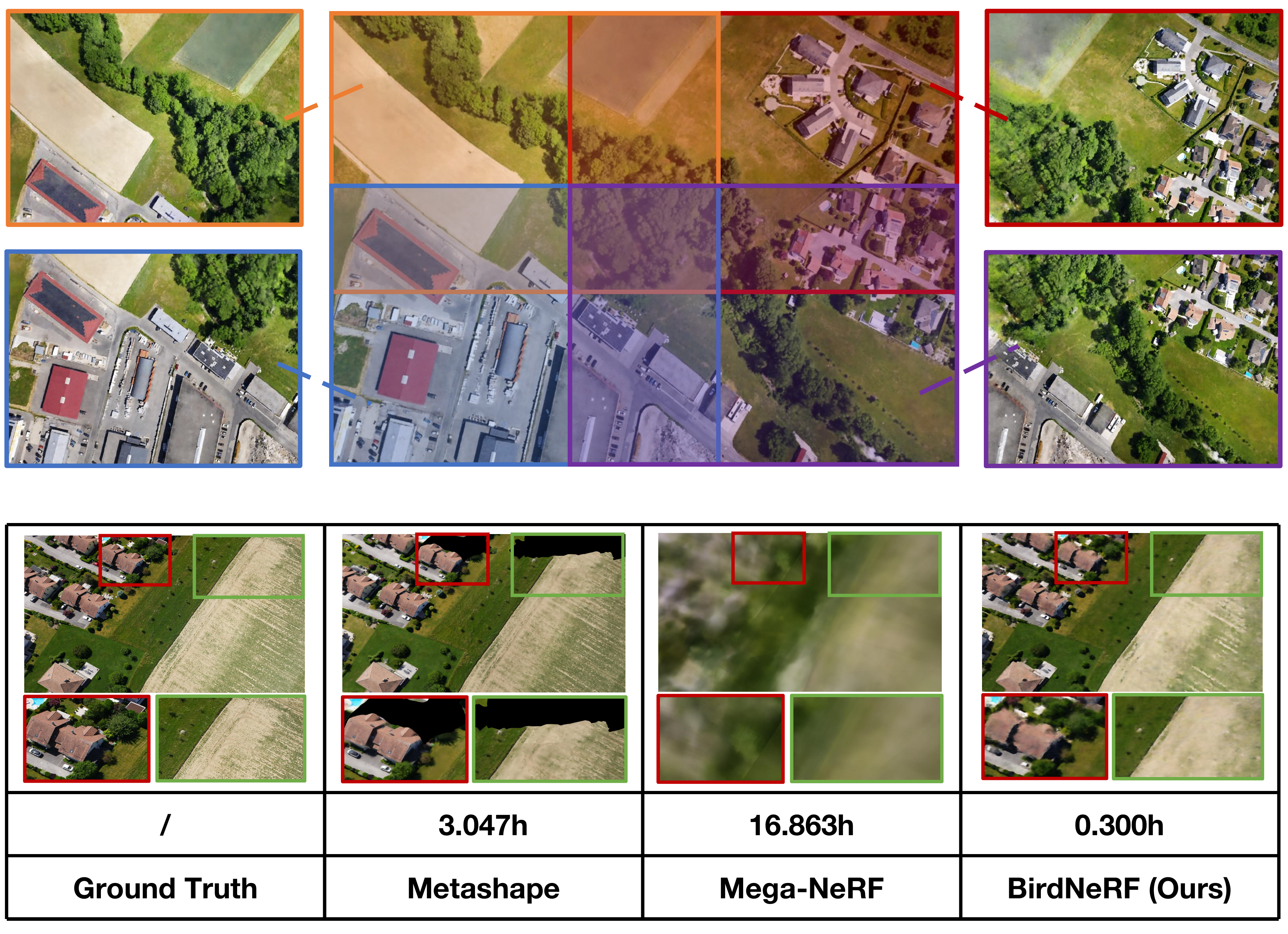}
    \caption{Illustration of modular scene training, along with performance and time comparisons on the IZAA dataset (comprising 1469 images). We demonstrate an approximately 10x speed improvement over traditional Metashape software. Moreover, when compared to current large-scale reconstruction approaches using deep learning, our method exhibits an approximately 56x enhancement in speed.} 
    \label{fig_intro}
\end{figure}

\IEEEPARstart{L}{arge-scale} 3D reconstruction on a city-wide level is an intrinsically active and significant task in areas of photogrammetry and remote sensing. This process revolves around constructing detailed and precise 3D models of entire cities utilizing an array of data sources, including, but not limited to, aerial or satellite images, LiDAR data, and street-level imagery. The expeditious advancements in aerial surveying technology have simplified and made cost-effective the procurement of high-resolution images. Consequently, image-based 3D reconstruction has emerged as an affluent and promising domain of study, encompassing manifold applications.

3D urban models find extensive application across various fields, providing a significant impetus to them. For example, urban development reaps the benefits of these models as they facilitate simulations and visual demonstrations of various scenarios, such as the erection of new edifices, the implications of transportation projects, and the layout of public venues, thereby aiding informed decision-making\cite{ref1}. In the sphere of navigation, 3D urban reconstructions contribute to the creation of accurate and exhaustive maps to augment the precision of GPS devices and mobile applications\cite{ref2}. Moreover, this technology forms the bedrock of augmented reality applications providing real-time overlaid directions on real world views. Virtual tourism thrives with extensive 3D reconstructions, as these enable individuals to virtually explore cities before actually visiting them and provide access to distant or otherwise difficult-to-reach areas\cite{ref3}. For the real estate industry, such reconstructions afford potential buyers a lucid and intuitive understanding of a property's environs, facilitating property valuation\cite{ref4}. 3D reconstructions hasten rescue operations, damage assessments, and post-disaster reconstruction planning during disaster management\cite{ref5}. Lastly, the domain of historical preservation considerably benefits from 3D reconstructions by aiding research, cultural heritage preservation, and the creation of virtual reconstructions for lost or damaged historical sites\cite{ref6}. These varied applications underscore the importance of 3D city modeling across diverse sectors.
\IEEEpubidadjcol  

Existing image-based 3D reconstruction techniques form two broad categories: traditional geometry-based methods and neural network-based methods. Geometry-based methods entail a two-step process composed primarily of Structure-from-Motion (SfM) and Multiple Views Stereo (MVS) \cite{ref7}. SfM estimates camera poses and sparse 3D points from input images \cite{ref8}, while MVS refines point clouds and builds a dense 3D model \cite{ref9}. Contrarily, neural network-based methods, epitomized by Neural Radiance Fields (NeRF) \cite{ref10}, signify a revolutionary progression in 3D reconstruction. NeRF employs neural networks to implicitly represent\cite{ref11,ref12} three-dimensional scene data by training network parameters grounded on input images and corresponding camera poses. It also depicts the capability to generate novel viewpoint images. More recent developments in NeRF research, such as Instant Neural Graphics Primitives (Instant-NGP) \cite{ref13}, accentuate the rapidly evolving terrain of this field. Nonetheless, large-scale urban reconstruction methods currently in use grapple with three main challenges: 
\begin{enumerate}
\item{ {\bf Slow Rendering with Large Models:} Large-scale reconstruction processes are inherently time-consuming. As the demand for real-time or near real-time applications, such as navigation and disaster management, intensifies, research is needed into faster and more efficient large-scale 3D reconstruction techniques.}
\item{ {\bf Computational Demands:} Large-scale reconstruction involves handling and processing vast datasets, often exceeding the memory capacity of a single GPU. This can result in slow processing times, out-of-memory errors, and other performance-related concerns, posing challenges to users or researchers with limited memory resources.}
\item{ {\bf Artifacts and Low Visual Fidelity:} Traditional geometry-based 3D reconstruction methodologies often grapple with challenges related to inaccurate camera pose estimation and limitations in model capacity. These issues manifest as artifacts and gaps in the reconstruction, leading to suboptimal visual fidelity.}
\end{enumerate}

Propelled by the intricate challenges inherent in large-scale scene reconstruction using aerial imagery, especially employing NeRF, we present BirdNeRF in this study, a dedicated adaptation of NeRF designed for reconstructions of large-scale aerial scenes. The proposed method of BirdNeRF chiefly incorporates spatial decomposition of camera distribution, followed by the modular training of smaller scenes (Fig.~\ref{fig_intro}), and finally generates images from novel viewpoints through our unique projection-guided view re-rendering strategy. BirdNeRF manages not only to enable reconstruction based on extensive aerial survey image inputs but also to guarantee excellent standards of reconstruction quality and speed of modeling. Fig.~\ref{fig_intro} demonstrates a comparative analysis of the quality of reconstruction between our proposed BirdNeRF and several other large-scale reconstruction methods, along with a comparison of training durations based on a dataset comprising 1469 images. The results depict significant improvements over previous solutions, proof of the efficacy of BirdNeRF in effectively addressing the challenges identified.

\subsection{Related Work and Motivations}
{\bf Structure-from-Motion and Multi-View-Stereo.} Structure-from-Motion (SfM)\cite{ref14} and Multi-View Stereo (MVS)\cite{ref9} combine to form a powerful pipeline for 3D reconstruction\cite{ref7}. SfM aims to recover camera poses and a sparse 3D structure of the scene. Conversely, MVS focuses on creating a dense 3D representation by estimating the depth or disparity of each pixel in the images, culminating in a dense point cloud. Often, MVS goes beyond mere dense reconstruction and incorporates surface reconstruction methods\cite{ref15,ref16}, resulting in either a mesh or continuous surface representation derived from the dense point cloud.

Prominent SfM and MVS techniques such as VisualSfM\cite{ref17}, COLMAP\cite{ref8}, and OpenMVG\cite{ref18} have made significant strides. However, these approaches can be hampered by scalability issues, slow processing speeds, and visual shortcomings, including holes, texture blending, and distortion\cite{ref19}. Insufficient information and inadequate image coverage in certain regions can lead to the problem of holes. Errors in camera calibration, image noise, and inaccurate feature matching can result in texture blending, visual distortions, and so on. Various post-processing techniques, such as hole filling, texture blending corrections and mesh refinement, are required to mitigate these challenges\cite{ref20}. Consequently, there's a concerted effort in ongoing research to enhance the accuracy, efficiency, and scalability of SfM and MVS algorithms, particularly for substantial reconstruction tasks.

{\bf Neural Radiance Fields.} Contrastingly, NeRF~\cite{ref10} employs a deep neural network to model the volumetric scene as a continuous function, bypassing the need for explicit geometrical or point-based representation. NeRF learns from a scene to predict sites appearance and density at any given 3D point, producing high-quality 3D reconstructions and novel views albeit with high computational demands.

Several extensions of naive NeRF have been developed to enhance this method. For instance, Instant-NGP~\cite{ref13} uses a hash encoding to accelerate the process remarkably, rendering it the fastest NeRF method currently. Similarly, NeRF++~\cite{ref21} and Mip-NeRF 360~\cite{ref22} have been fashioned specifically for unbounded scenes. DrRF~\cite{ref23} partitions the scene using spatial Voronoi and renders each image part independently, accelerating rendering by three times compared to NeRF~\cite{ref5}. KiloNeRF~\cite{ref24} partitions the scene and assigns it to thousands of small networks for collective training, which speeds up the inference process to a certain extent. However, DeRF and KiloNeRF need an extra expensive initialization~\cite{ref5}. The PixelNeRF~\cite{ref25} optimizes the NeRF model training by leveraging prior information from image features, facilitating rapid model reconstruction from sparse inputs. Nevertheless, none of these methods is suited for city-level reconstruction.

{\bf Commercial Reconstruction Software.} Pix4D Mapper~\cite{ref26} and Agisoft Metashape~\cite{ref27} are both well-known and widely used commercial software packages providing comprehensive solutions for photogrammetry and 3D reconstruction. They offer robust and reliable ways to process aerial and terrestrial images, generating accurate and detailed 3D models, point clouds, orthomosaics, and digital surface models. Benefiting from well-established algorithms for camera calibration, feature matching, dense point cloud generation, and mesh reconstruction, they can deliver an exceptional reconstruction.

Despite their capabilities, these software tools have inherent limitations and challenges. Photogrammetry and 3D reconstruction processes are computationally intensive, especially when the task involves large datasets or highly complex scenes. As such, Pix4D Mapper and Agisoft Metashape both demand a significant amount of computational resources, including processing power, memory, and ample storage. Users attempt to run this software on lower-end machines or systems with limited resources may encounter sluggish processing times or reduced performance.

{\bf Large Scale Reconstruction.} Over the past few decades, wide-ranging efforts have been made towards achieving large-scale reconstruction. Studies such as~\cite{ref28,ref29} focus on employing parallelism in the reconstruction process. A substantial breakthrough in large-scale reconstruction tasks has been accomplished by~\cite{ref30,ref31,ref32} through the application of SfM and MVS methods. For instance,~\cite{ref30} introduces a divide-and-conquer framework for handling large scale global SfM, while \cite{ref31} proposes a distributed method to address global bundle adjustment for colossal scale SfM computations. On the other hand,~\cite{ref32} employs surface-segmentation-based camera clustering to achieve the decomposition of large-scale MVS. These novel approaches provid significant inspiration for our work.

Several examples of large-scale reconstruction works, such as~\cite{ref5,ref19,ref33}, are based on NeRF. Despite Mega-NeRF~\cite{ref5} achieving large-scale reconstruction on a single GPU, it suffers from extremely long training times due to its dependency on the original NeRF implementation. Block-NeRF~\cite{ref19} primarily uses images captured by vehicle-mounted cameras and decomposes the scene into distinct spatial units, each corresponding to a fixed city block. However, this method demands significant training resources. Furthermore, BungeeNeRF~\cite{ref33} models diverse multi-scale scenes using multiple data sources. Although these methods have achieved their goal of large-scale reconstruction, they have done so at the cost of extensive training resources, while also not effectively addressing the challenge of limited GPU memory in practical applications.

{\bf Motivations.} From the above discussion, it becomes evident that a fast, high-quality, large-scale 3D reconstruction methodology that operates within limited resource constraints is yet to be developed. The prevailing methods, especially those grounded in NeRF for 3D modelling, impose considerable demands on the training resources. This high computational burden, along with elongated training durations, often results in constrained model capacity, leading to visual distortions such as artifacts, voids, and blurring. A quick and high-quality modelling process is imperative, especially in vital fields such as urban planning and disaster response, making exploration of expedited, high-quality, large-scale 3D reconstruction methodologies that work within limited memory constraints both pertinent and necessary. These efforts hold significant potential for catering to the growing and changing needs of such crucial applications.

\subsection{Contributions}
In response to the substantial challenge of enabling swift, high-quality, large-scale 3D reconstruction within the bounds of limited memory resources, we present BirdNeRF. The process begins with a spatial decomposition, using the spatial distribution of cameras to discern separate clusters, thereby segmenting the training scenes. Each resulting sub-scene is then trained independently. In the final stage, our unique projection-guided novel view re-rendering strategy is utilized to register and align query cameras, facilitating the rendering of the requested query viewpoints. This methodical approach, integrating spatial decomposition, independent training, and a customized re-rendering strategy, optimizes large-scale 3D reconstruction in resource-limited settings. We conducted extensive experiments to evaluate our approach, providing both qualitative and quantitative insights into its effectiveness. The outcomes highlight its distinct advantages in terms of both modeling time and the quality of the rendered images. The primary contributions of this study can be summarized as follows:

\begin{enumerate}
\item{{\bf Unprecedented speed in large-scale reconstruction:} We introduce a pipeline for large-scale reconstruction that is unparalleled in its speed. Our method is capable of reconstructing scenes of up to 1 square kilometer in approximately half an hour, eclipsing the speeds of commercial software like Metashape by a factor of ten or more, and outperforming current deep learning approaches by more than fifty times. This speed advantage only increases for larger datasets.}
\item{{\bf Adaptability to GPU memory constraints:} Our methodology is marked by its adaptability to a variety of GPU memory resources. Through the use of a spatial decomposition strategy based on camera distribution, we are able to manage large-scale reconstructions effectively within the constraints of limited GPU memory. This adaptability makes our method versatile and scalable, demonstrating its applicability across a range of hardware setups.}
\item{{\bf Innovative re-rendering strategy for high-quality results:} We introduce an innovative projection-guided novel view re-rendering strategy that ensures precise registration and query of cameras during the rendering process. This strategy meticulously brings the relevant sub-models into play for rendering output, guaranteeing an accurate and efficient combination of rendered images from various viewpoints across scenes of all sizes.}
\end{enumerate}

\section{Methodology}
Representing and reconstructing large-scale scenes, such as those present in aerial imagery, poses considerable challenges due to the inherent scalability limitations of training a single NeRF. In addressing these challenges, we propose BirdNeRF, a method characterized by decomposing the environment into a series of NeRFs, each trained individually based on the bird view field of view (FOV). During the inference phase, the novel view is rendered by aggregating the outputs from these disparate NeRFs. This strategy, which we term the "split-unite paradigm", successfully circumvents the limitations of model capacity that have stymied previous NeRF research, enabling efficient reconstruction of expansive scenes even with limited computational resources.
    
\begin{figure*}[htbp] 
    \centering
    \includegraphics[width=1\linewidth]{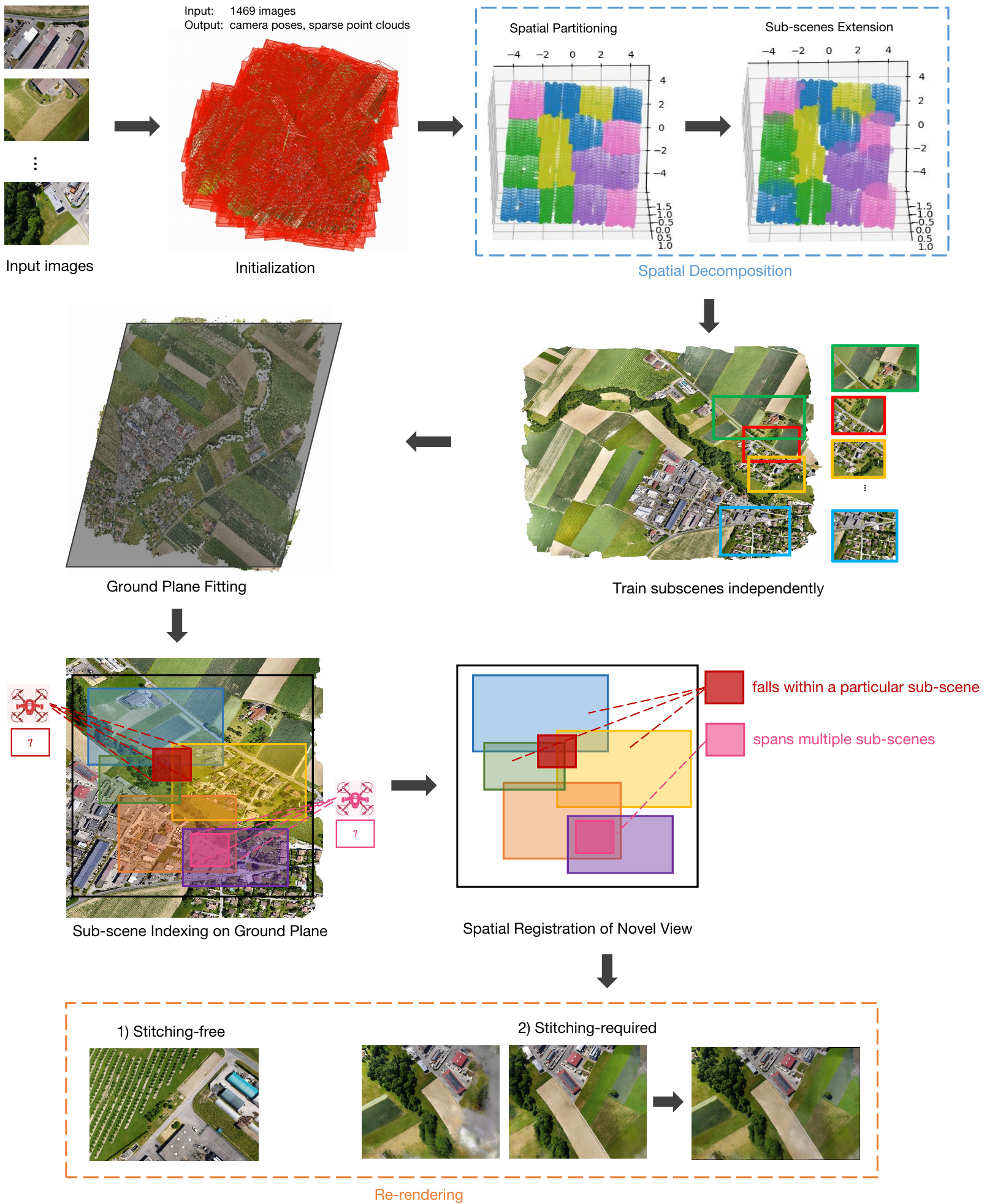} 
    \caption{The BirdNeRF pipeline is initiated by the preprocessing phase \ref{sec:init}, where input images are processed to obtain camera positions and coefficient point clouds. Spatial decomposition follows in section \ref{sec:spa_dcp}, categorizing cameras into clusters. For each cluster, associated images facilitate independent training mentioned in section \ref{sec:train}, creating multiple sub-scenes. The novel projection-guided view re-rendering strategy described in section \ref{sec:projection} synthesizes the final rendering images.
    }
    \label{pipline}
\end{figure*}

As depicted in Fig.~\ref{pipline}, BirdNeRF encompasses two principal phases:
(1) spatial decomposition, which involves dividing the scene into manageable, cluster-based segments, and 
(2) projection-guided novel view re-rendering, which reunites the independently processed segments to render the desired viewpoint. \

\subsection{Background}
BirdNeRF is fundamentally grounded in the original NeRF~\cite{ref10} methodology and its subsequent advancement, Instant-NGP~\cite{ref13}. Herein, we provide a synopsis of these foundational methods, while detailed expositions can be found in the respective source papers.\\

\noindent $\bullet$ \textbf{NeRF overview.} NeRF is a transformative model that introduces a fresh paradigm to scene representation and view synthesis, enabling the creation of photorealistic 3D scenes from a set of 2D images. NeRF operates by associating each pixel in an image with a corresponding ray in 3D space, estimating color and density along these rays, and adjusting network parameters to minimize the difference between observed and predicted colors. Once trained, NeRF can be leveraged to synthesize novel views by projecting rays from new camera positions and integrating the color and density estimates to produce lifelike images from viewpoints not captured in the original dataset.

To elucidate, let's consider an image pixel at the camera center with specific pixel coordinates. For this pixel, NeRF constructs a ray $\boldsymbol{r}(t) = \boldsymbol{o} + t\boldsymbol{d}$ in 3D space. The points along this ray are represented by their position $\boldsymbol{x}^i = [x, y, z]$ and direction $\boldsymbol{d}^i = [d1, d2, d3]$. Positional encoding (Eq.~\eqref{eq-b1}) is then applied to these coordinates, enabling NeRF to capture higher frequency details in the scene. These encoded coordinates are then fed to the NeRF model, a Multilayer Perceptron (MLP), which outputs the color $c^{i} = [r, g, b]$ and density $\sigma^{i}$ for each point.

\begin{equation}
   \begin{split}
    \label{eq-b1}
    \gamma(p)=(\sin (2^{0}\pi  p), \cos (2^{0}\pi  p),...,\\
    \sin(2^{L-1}\pi  p), \cos(2^{L-1}\pi  p) )
   \end{split}
\end{equation}
where L is the number of levels of positional encoding, NeRF sets L = 10 for $\gamma(\boldsymbol{x}^i)$ and L = 4 for $\gamma(\boldsymbol{d}^i)$.\\

The volume rendering process (Eq.~\eqref{eq-b2}) integrates these color and density estimates to calculate the final pixel color $\hat{C}(\boldsymbol{r})$ for the ray $\boldsymbol{r}$.
\begin{equation}
    \begin{split}
    \label{eq-b2}
    \hat{C}(\boldsymbol{r})=\sum_{i=1}^{N} T_{i}(1-exp(-\sigma_{i}\delta_{i}))c_i, \\
    where \  T_i=exp(-\sum_{j=1}^{i-1} \sigma_{j}\delta_{j})
    \end{split}
\end{equation}
where $\delta_{i}$ is the distance between samples $p_i$ and
$p_{i+1}$.  \\

The loss function $L$ (Eq.~\eqref{eq-b3}) is computed by comparing the rendered color value to the original image color value to supervise the model training. By training the model, an implicit representation of the 3D scene is obtained, allowing for rendering arbitrary camera views from different viewpoints.
\begin{equation}
    \begin{split}
    \label{eq-b3}
    L=\sum _{r\in \Re} \left \| C(\boldsymbol{r})-\hat{C}(\boldsymbol{r}) \right \| ^2
    \end{split}
\end{equation}
where $\Re$ is the set of rays in each batch, $C(\boldsymbol{r})$ is the ground truth and $\hat{C}(\boldsymbol{r})$ is the predicted RGB colors for ray $\boldsymbol{r}$ respectively.\\

\noindent $\bullet$ \textbf{Instant-NGP.} Instant-NGP~\cite{ref13} is a high-performance neural radiance field method developed by NVIDIA. It distinguishes itself by implementing a multi-resolution hash encoding approach, which accelerates the training process. The method also incorporates an enhanced multi-resolution hash table, designed to accommodate trainable feature vectors and reduce the overall model size. Instant-NGP is a fully integrated system, offering a seamless and efficient deployment for rapid training and real-time rendering of detailed scenes.

\subsection{Initialization} \label{sec:init}
Our methodology begins with the utilization of the classical Multi-View Stereo (MVS) software COLMAP~\cite{ref8} on a dataset of aerial images. This step involves computing camera poses and generating a sparse point cloud for the scene. The camera model is set as a pinhole camera model, using uniform internal parameters for feature extraction and matching. An incremental reconstruction approach combined with camera pose estimation is applied to obtain sparse point clouds representing the scene and pose information for all cameras in the dataset.

\subsection{Spatial decomposition}\label{sec:spa_dcp}

\noindent $\bullet$ \textbf{Spatial partitioning.}
Once the camera poses and sparse point clouds are obtained from the initialization phase in Section \ref{sec:init}, an initial partitioning step is performed based on the spatial coordinates of the cameras. This is done using the K-Means clustering algorithm~\cite{ref34}. The optimal number of clusters, denoted as K, is determined based on the available GPU memory size. Each cluster represents a sub-scene in the dataset.\\

\begin{figure}[htbp] 
    \centering
    \includegraphics[width=1\linewidth]{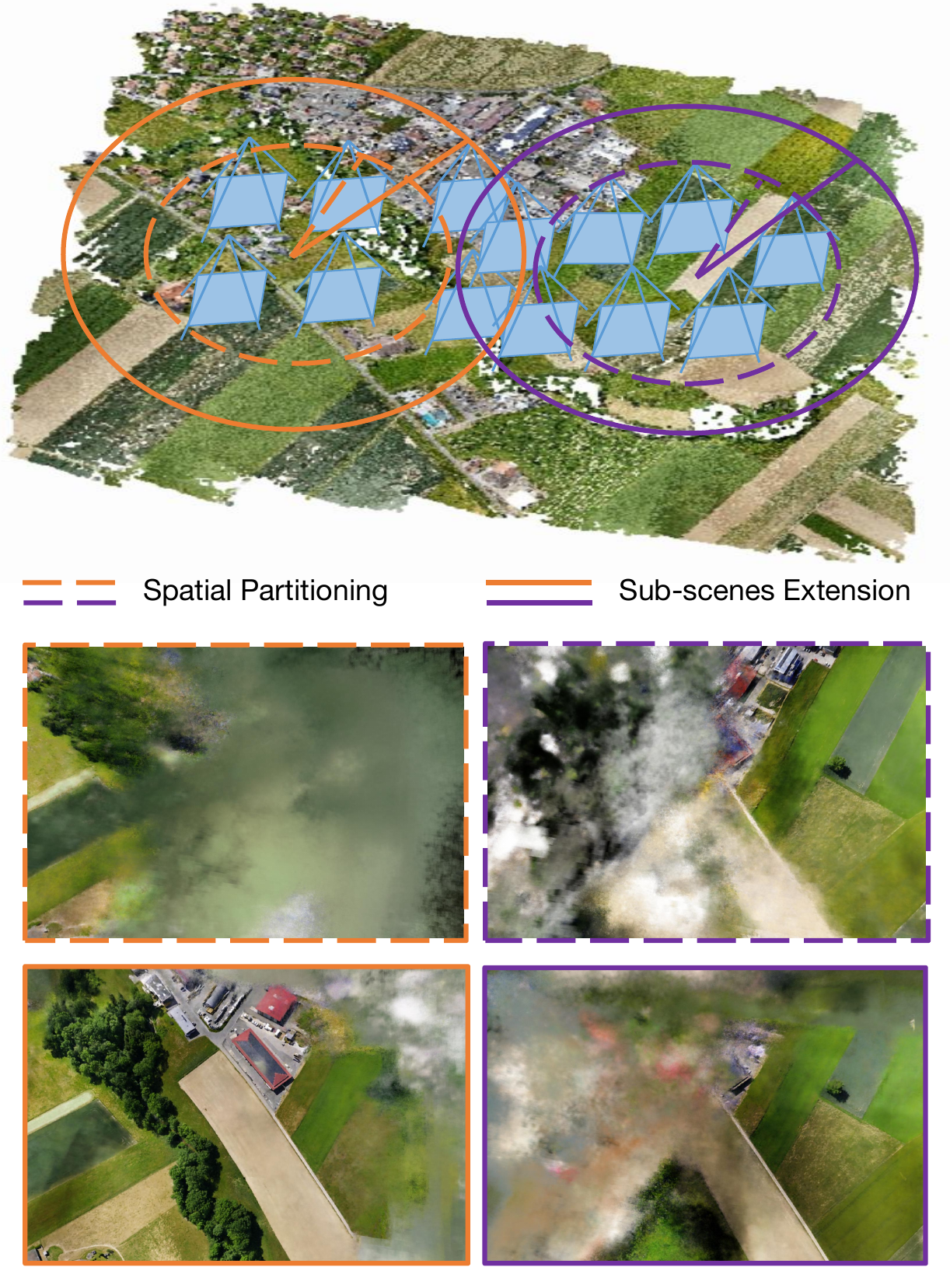} 
    \caption{Sub-scenes extension. The strategic expansion of sub-scenes enhances scene overlap, thereby elevating the success rate of post-image registration in our proposed approach.}
    \label{fig_0}
\end{figure}

\noindent $\bullet$ \textbf{Sub-scenes extension.} 
We extend the sub-scenes to ensure a defined degree of overlap, enhancing the efficacy of image registration, as depicted in Fig.~\ref{fig_0}. Concretely, we introduce an expanded threshold denoted as $\sigma$ for the number of cameras, guiding the augmentation of each scene to guarantee a predetermined level of overlap within the partitioned sub-scenes. The computation of intra-cluster distances, represented by $d_{k}$ for each cluster, facilitates the identification of the maximum intra-cluster distance. This maximum distance is then multiplied by the designated threshold to determine the new cluster diameter. Subsequently, utilizing the preceding cluster centroids as centers, we search for cameras within the range of the recalibrated diameter and incorporate them into the updated clusters. It is noteworthy that a maximum limit on the number of images per sub-scene is imposed to ensure seamless training within the allocated GPU memory resources. This stage signifies the completion of camera division, denoting the subdivision of the scene into distinct sub-scenes. A comprehensive algorithmic depiction is presented in Algorithm~\ref{alg:alg1}.\\

\begin{algorithm}[htbp]
    \caption{Spatial decomposition algorithm.}\label{alg:alg1}
    \begin{algorithmic}
        \STATE 
        \STATE \textbf{Input:} 
        \STATE \hspace{0.5cm}Maximum number of cameras per subscene, \textbf{maxN}. 
        \STATE \hspace{0.5cm}Total number of cameras in a dataset, \textbf{N}.
        \STATE \hspace{0.5cm}The expanded threshold, $ \boldsymbol{\sigma} $.
        \STATE \textbf{Process:}
        \STATE \hspace{0.5cm} K ← N / maxN 
        \STATE \hspace{0.5cm} Labels, Centers ← KMeans(K)
        \STATE \hspace{0.5cm} \textbf{for} $k \in \{1,...,K\}$ \textbf{do}
        \STATE \hspace{0.8cm} \textbf{for} $i \in \{1,...,maxN\}$ \textbf{do}
        \STATE \hspace{1cm}$ d_{k} \gets \textsc{max}$ (EuclideanDistances($T_i,Center_k$))
        \STATE \hspace{0.8cm} \textbf{endfor}
        \STATE \hspace{0.5cm} \textbf{endfor}
        \STATE \hspace{0.5cm} \textbf{for} $k \in \{1,...,K\}$ \textbf{do}
        \STATE \hspace{0.65cm}$ d_{k}' \gets \sigma * d_k$
        \STATE \hspace{0.5cm} \textbf{endfor}
        
        \STATE \hspace{0.5cm} \textbf{for} $k \in \{1,...,K\}$ \textbf{do}
        \STATE \hspace{0.8cm} \textbf{for} $i \in \{1,...,N\}$ \textbf{do}
        \STATE \hspace{0.8cm} \textbf{if} EuclideanDistances($T_i,Center_k$))$ < d_{k}'$ :
        \STATE \hspace{1.3cm}$ Scene_{k} \gets T_i $
        \STATE \hspace{0.8cm} \textbf{endfor}
        \STATE \hspace{0.5cm} \textbf{endfor}
     
        \STATE \textbf{Output:}
        \STATE \hspace{0.5cm} The subscene partition obtained through expanded clustering, $ \boldsymbol{Scene_{k} (k=1,...,K)}$.
        
    \end{algorithmic}\label{alg1}
\end{algorithm}

\subsection{Individual training}\label{sec:train}
After the spatial decomposition process, we obtain separate camera clusters where the camera parameters and corresponding images within each cluster form the training data for the respective sub-scenes. Independent training is performed for each sub-scene using Instant-NGP as the base training model. Once the training is completed, the resulting model parameters are stored offline on disk for future use in the re-rendering process. It is important to note that the models are stored as network parameters, which occupies much less disk space compared to alternative representations such as point clouds or grids. This efficient storage of the model parameters allows for easier access and retrieval when needed.

\begin{figure*}[htbp] 
    \centering
    \includegraphics[width=1\linewidth]{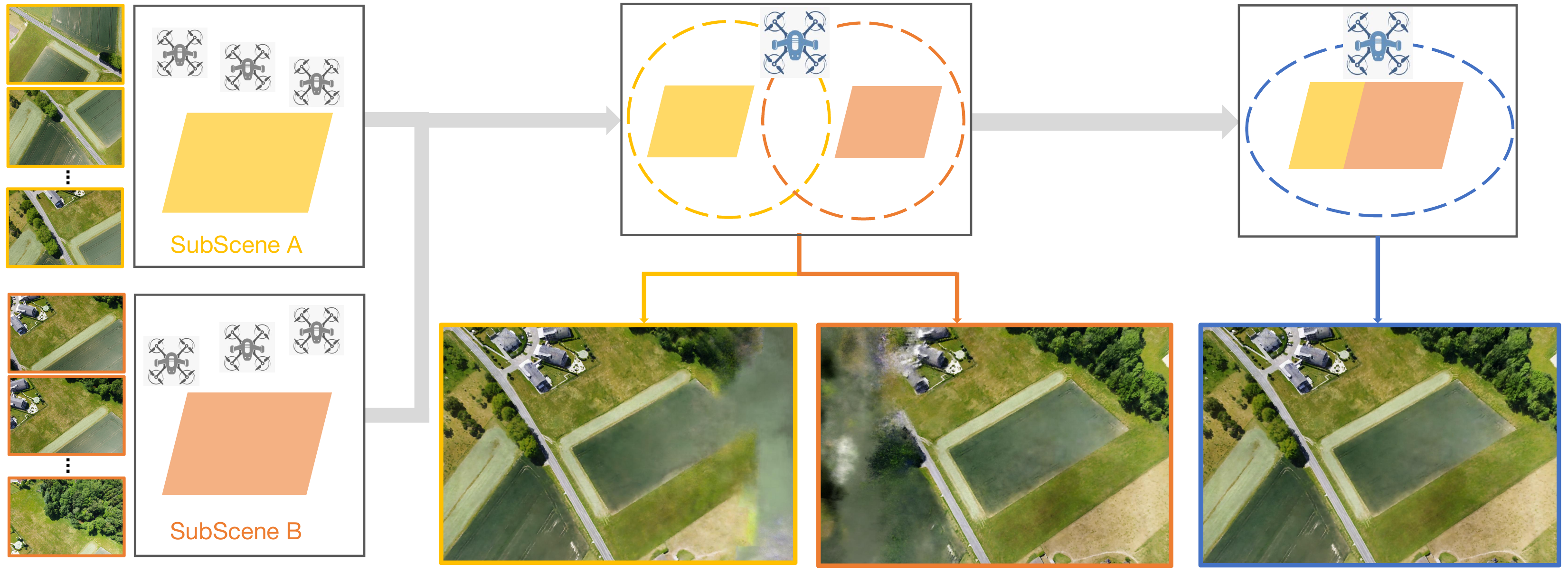} 
    \caption{Projection-guided novel view re-rendering. Beginning with independently constructed input NeRFs, namely Sub-scene A and Sub-scene B, we perform image rendering from novel viewpoints. Then, employing a sequence of image stitching and fusion techniques, we achieve higher-quality re-rendering results.}
    \label{demo}
\end{figure*}

\subsection{Projection-guided novel view re-rendering}\label{sec:projection}
In order to effectively query target scenes, we employ a set of methods to optimize the novel view fusion stage, as shown in Fig.~\ref{demo}. \\

\noindent $\bullet$ \textbf{Ground plane fitting.} Prior to the subsequent procedure, the ground plane parameters are initially determined using the Least Squares~\cite{ref35} approach.\\

\noindent $\bullet$ \textbf{Sub-scene indexing on ground plane.} We employ a methodology involving the projection of the four corner points of an image onto the ground plane, a plane fitted through the sparse point cloud. This technique facilitates the determination of the scene extent corresponding to the image captured by the camera.
  
In this study, the impact of camera type on our results is negligible. Consequently, we default to assuming our method is based on the pinhole camera model. For a pinhole camera, the coordinates of the four corner points of the $i-th$ image in the camera coordinate system are denoted as
\begin{equation}
\begin{split}
p_1^i = [0-cx,0-cy,f_i]\\
p_2^i = [w-cx,0-cy,f_i]\\
p_3^i = [w-cx,h-cy,f_i]\\
p_4^i = [0-cx,h-cy,f_i]
\end{split}
\end{equation}

The spatial coordinates of the optical center associated with the $i-th$ camera, corresponding to a given image, are denoted as
\begin{equation}
\begin{split}
o^i = [0,0,0]
\end{split}
\end{equation}

Utilizing the pinhole camera model, we transform the coordinates of the four corner points and optical center from the camera coordinate system to the world coordinate system, as indicated by Eq.~\eqref{eq1}. 

\begin{equation}\label{eq1}
    P^i_k = R^i \cdot p^i_k + T^i
\end{equation}

Here, $P^i_k$ signifies the coordinates of the $k-th$ corner point in the world coordinate system of the $i-th$ camera, while $p^i_k$ denotes the coordinates of the same point in the camera coordinate system of the $i-th$ camera. The parameters $R^i$ and $T^i$ characterize the camera pose with respect to the world coordinate system.

The conversion of the coordinates for the camera's optical center is likewise conducted using the aforementioned equation, as indicated by Eq.~\eqref{eq1-1}.

\begin{equation}\label{eq1-1}
    O^i = R^i \cdot o^i + T^i
\end{equation}

\begin{figure}[htbp]
    \centering
    \includegraphics[width=1\linewidth]{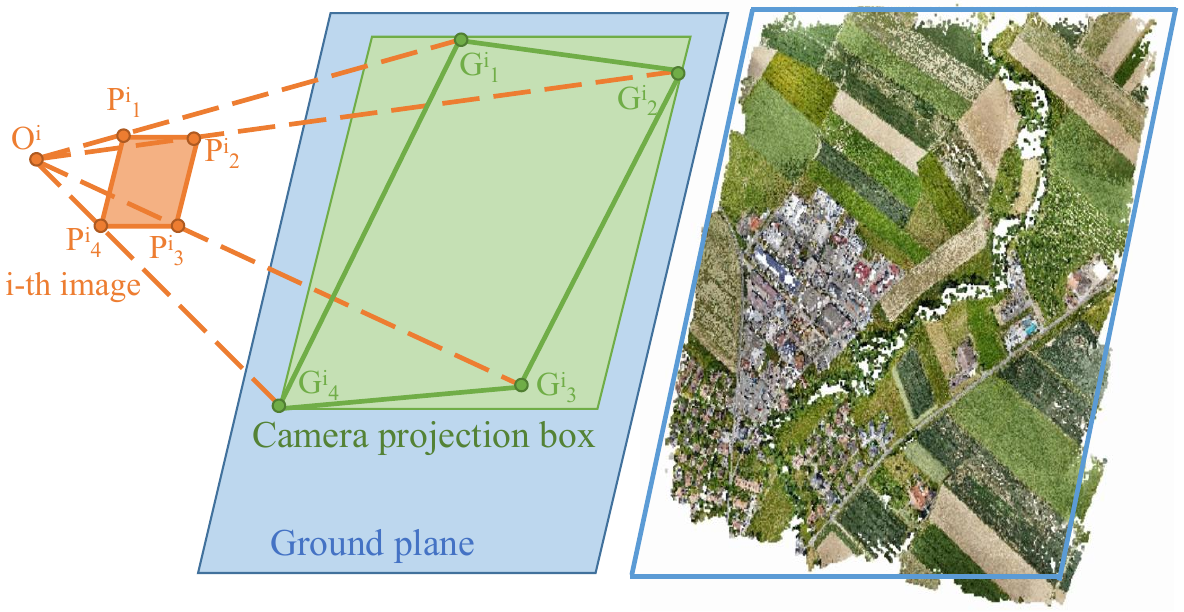}
    \caption{Ground plane fitting and pixel projection.}
    \label{fig_1}
\end{figure}

Following this, the intersection of line segments, formed by connecting the optical center and the corner points, with the ground plane is computed to derive the intersection points, as depicted in Fig.~\ref{fig_1}. The minimum rectangle that encompasses all such points is determined upon obtaining these intersection points. This rectangle effectively represents the extent of the scene captured by the $i-th$ camera, known as the camera projection box.

\begin{figure}[htbp]
    \centering
    \includegraphics[width=0.8\linewidth]{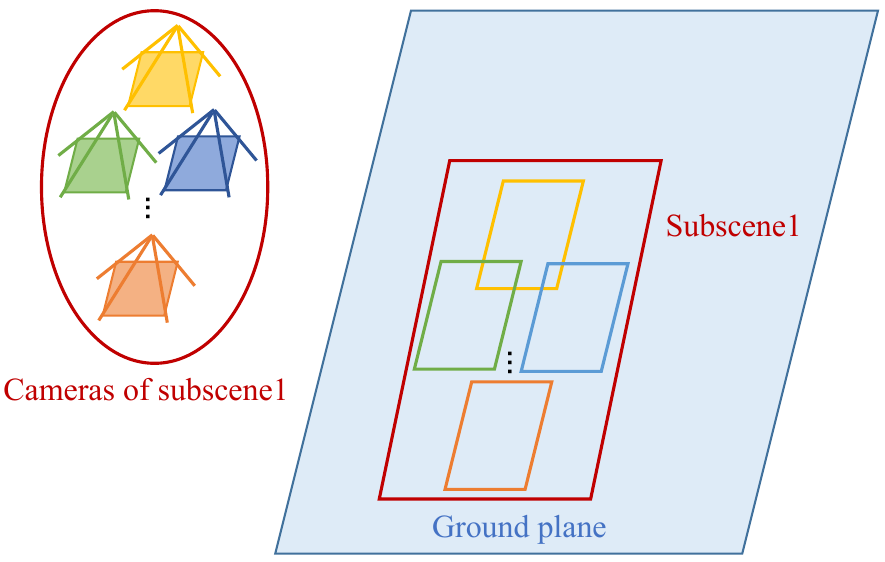}
    \caption{Sub-scene bounding Box. The projection boxes of all cameras within each divided sub-scene collectively form the projection box of the sub-scene range.}
    \label{fig_2}
\end{figure}

As detailed in Section~\ref{sec:spa_dcp}, we employ a spatial decomposition methodology to allocate each camera to multiple sub-scenes. Each sub-scene comprises several cameras, with its bounding box defined as the amalgamation of the individual bounding boxes of all cameras assigned to that specific sub-scene. Furthermore, a minimum bounding rectangle is computed to encapsulate the collective scene range captured by all cameras within the sub-scene. This resulting bounding rectangle serves as the scene bounding box for the sub-scene, as visually represented in Fig.~\ref{fig_2}.\\

\noindent $\bullet$ \textbf{Spatial registration of novel view.} As per the methodology employed in the preceding section, we compute the projection box of the queried camera onto the ground plane. Subsequently, we systematically iterate through all the sub-scene boxes to identify those that intersect with the projection box of the queried camera. These identified sub-scene bounding boxes are preserved as rendering schemes, guiding the subsequent rendering process for the queried image. This rendering process is executed by leveraging the trained models associated with each respective sub-scene.\\

\noindent $\bullet$ \textbf{Re-rendering.} Once we obtain the bounding boxes for each independently trained sub-scene, we can automatically generate a rendering strategy for the query camera that needs to be rendered.  Generally speaking, there are two situations when querying scenes (Fig.~\ref{fig_3}): 

\begin{figure*}[htbp]
    \centering
    \subfloat[Querry camera falls within a sub-scene]{\includegraphics[width=0.46\linewidth]{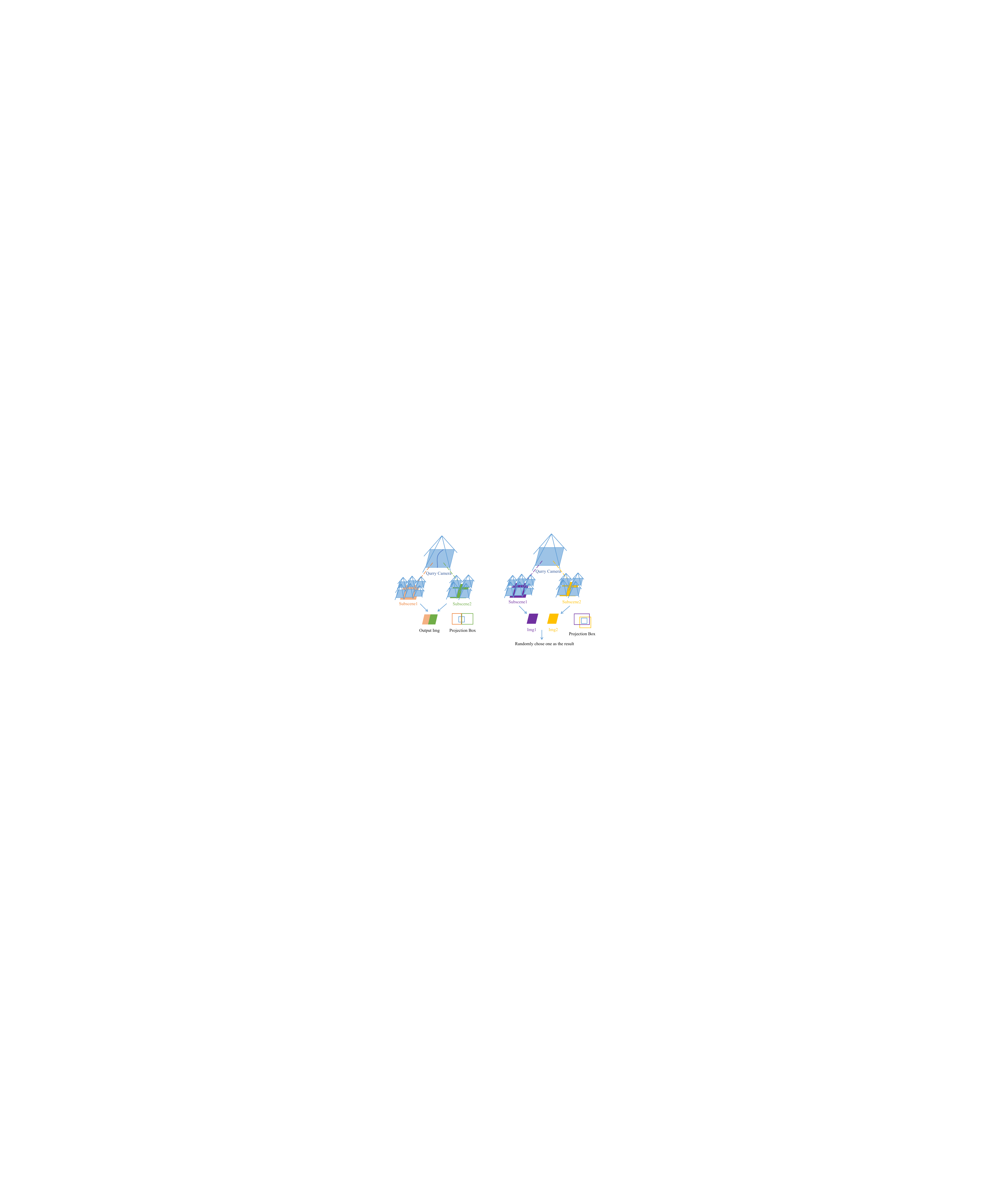}%
    \label{fig_first_case}} 
    \hfil
    \subfloat[Querry camera spans multiple sub-scenes]{\includegraphics[width=0.46\linewidth]{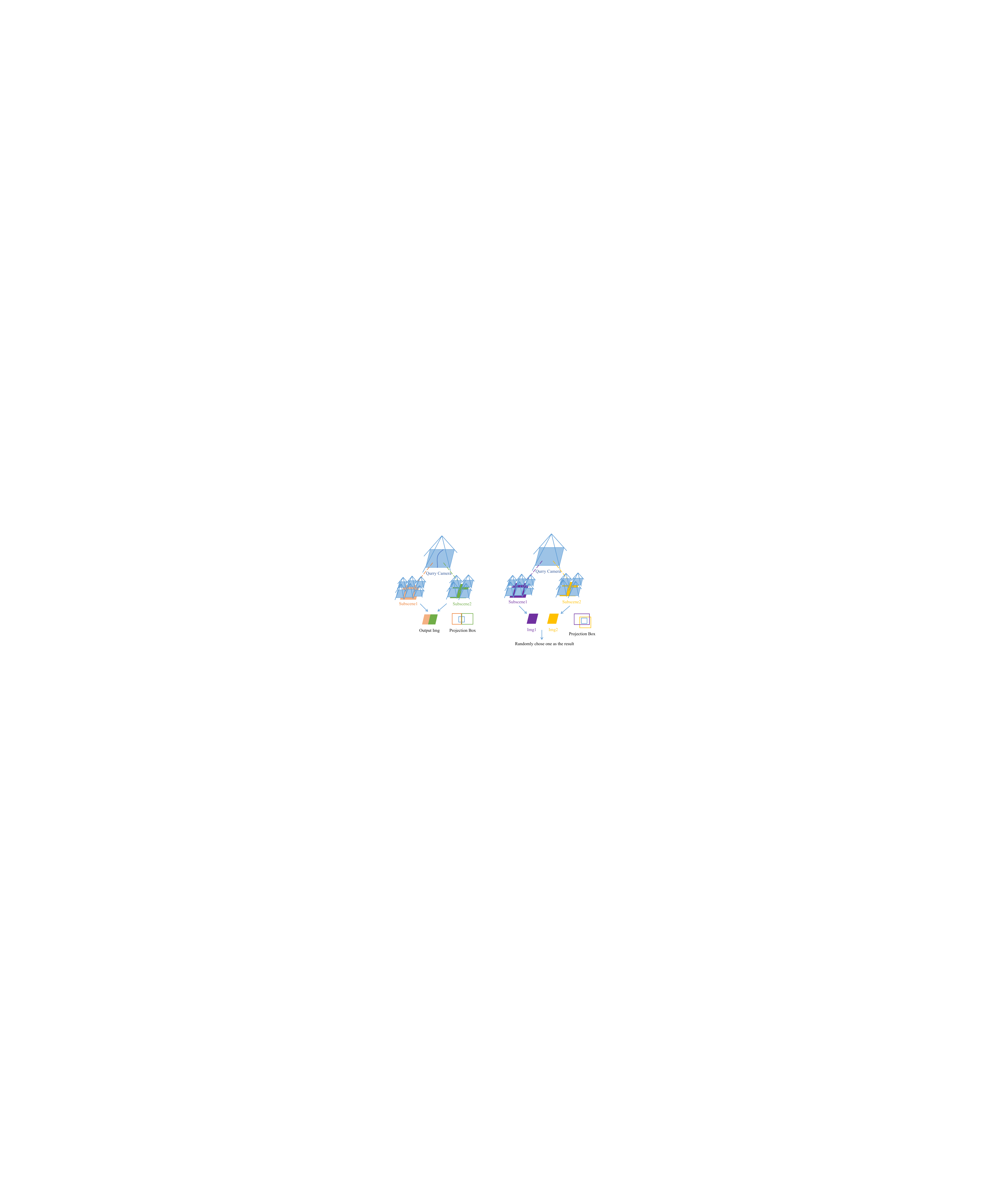}%
    \label{fig_second_case}} 
    \caption{Two types of query camera position distributions. When the query camera projection box is entirely contained within the bounding box of a specific sub-scene, the rendered image is directly output from that sub-scene. In cases where the query camera is positioned at the boundary of multiple sub-scenes, registration, and fusion of outputs from these sub-scenes are necessary to obtain the final result.}
    \label{fig_3}
\end{figure*}

\subsubsection{Stitching-free}
In situations where the retrieved scene lies entirely within a specific sub-scene, rendering can be performed using the network model of that sub-scene alone. Fig.~\ref{fig_first_case} illustrates an example where the identified scene matches the requirements for stitching-free rendering. In this case, the intersection of projection boxes is applied to the sub-scenes that encompass the queried camera's placement. Since the camera resides within a single sub-scene, a complete rendering can be achieved by using that sub-scene alone. Therefore, the resulting image can be generated from any of these relevant sub-scenes.

\begin{figure}[htbp] 
    \centering
    \includegraphics[width=1\linewidth]{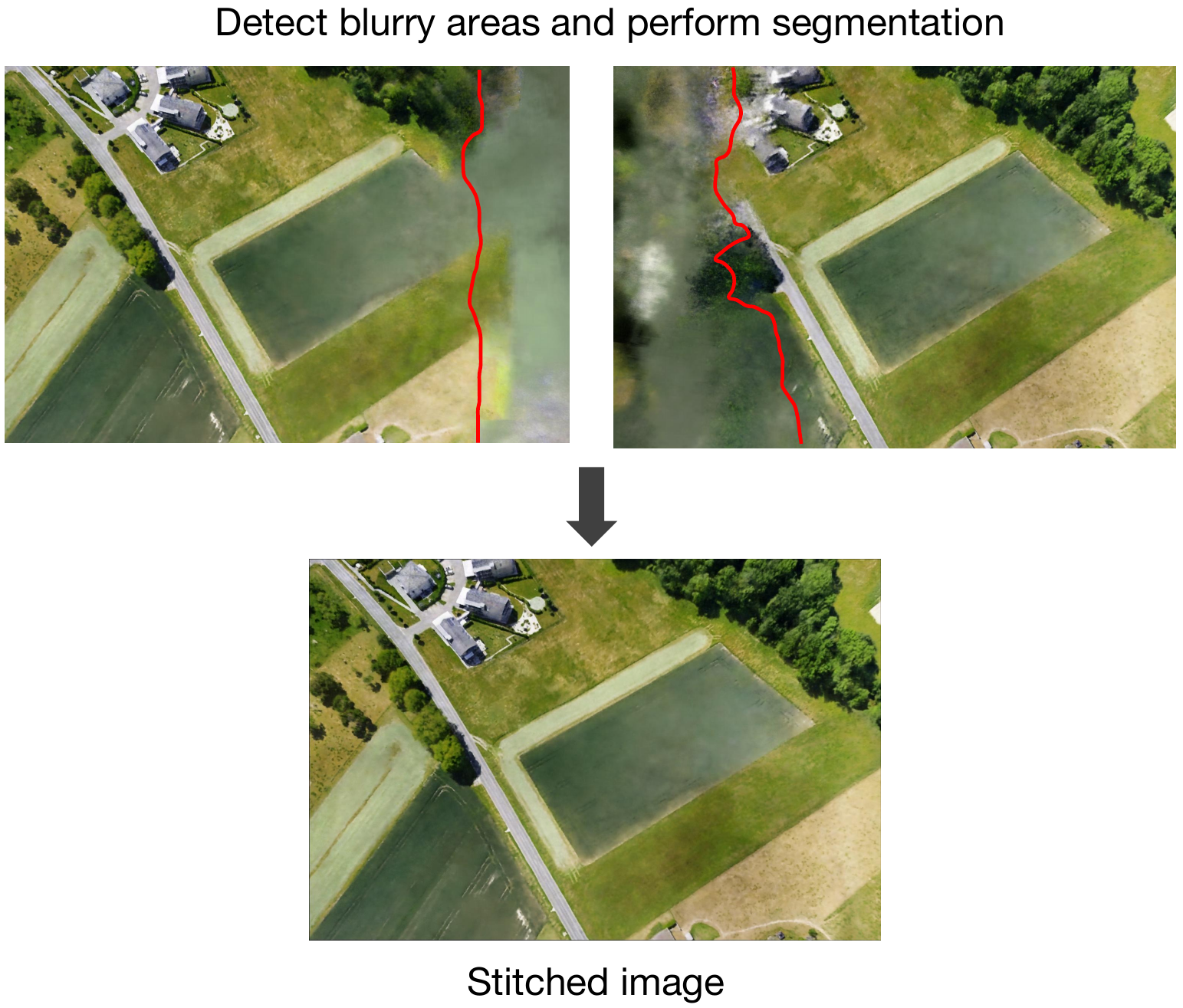} 
    \caption{Blur area detection and image stitching.}
    \label{blur}
\end{figure}

\subsubsection{Stitching-required}
In this case, the retrieved scene range does not fall entirely within a certain sub-scene range. We need to render multiple sub-scene network models containing the retrieved scene in order to render and stitch together an image containing the entire target scene range. As illustrated in Fig.~\ref{fig_second_case}, when the queried camera is situated at the boundaries of multiple scenes, the resulting rendered image requires collaborative composition from various sub-scenes, where each scene contributes to rendering a distinct portion of the image. Subsequent to invoking these sub-models for rendering and generating their respective images, we undertake the detection and segmentation of the blurred areas within each rendered image\cite{ref39}, as delineated in Fig.~\ref{blur}. Following the removal of blurred portions, we employ an image stitching algorithm\cite{ref36,ref37}, which integrates multiple panoramas, incorporates gain compensation, employs simple blending, and implements multi-band blending. This comprehensive approach amalgamates the partial images into a coherent new rendering result.

\section{Experiments}
\subsection{Implementation}
We run our method on a 12th Gen Intel(R) Core(TM) i9-12900KF with 32GB RAM. All the experiments in this paper are conducted on a single NVIDIA GeForce RTX 3090 GPU(24GB). We set the default number of cameras in a sub-scene to 90 in order to determine the initial clustering clusters (K value). Additionally, we set the maximum number of cameras in each partitioned sub-scene to 115 to ensure that our method runs without encountering GPU memory overflow. This value can be adjusted according to the GPU specifications of the experimental environment. We set the expanded threshold $\sigma$ to 1.1 when scaling the maximum intra-cluster distance for camera cluster augmentation. For the model training for each sub-scene, we set the number of training iterations to $5 \times 10^3$, while keeping the other parameters at their default values.

\subsection{Datasets}
Our experiments utilize aerial datasets from 8 distinct geographical regions, each varying in size and characteristics. The datasets are selected from both publicly available sources and those captured using the DJI Mavic Air 2 drone in Hexi University Town, Changsha. These datasets collectively cover urban, suburban, industrial, agricultural, and university campus environments, providing diverse scenes for comprehensive evaluation.\\

We select three real photogrammetry datasets from Pix4D's example projects\cite{ref38}:
\begin{itemize}
  \item{{\bf Urban area(UA).} UA represents a city dataset covering 0.0214 $km^2$, featuring 100 images with a resolution of 6000 x 4000 pixels. }
  \item{{\bf Suburban area(SA).} SA) comprises 188 images at a resolution of 5472 x 3648 pixels, covering an area of 0.041 $km^2$. }
  \item{{\bf Industrial zone and agriculture area(IZAA).} IZAA dataset consist of 1469 images at a resolution of 6000 x 4000 pixels, encompassing an extensive area of 1.154 $km^2$. This dataset includes diverse regions such as an industrial zone, suburban residential areas, and surrounding agricultural landscapes. }
\end{itemize}

Utilizing the DJI Mavic Air 2 drone, we conduct aerial surveys capturing five distinct regions within Hexi University Town, situated in Changsha, Hunan Province. These surveys result in the creation five datasets, varying in size from hundreds to thousands of images. Notably, the flight trajectories of our drone missions introduce additional complexities compared to the publicly available datasets previously mentioned.
\begin{itemize}
  \item{{\bf CSU1.} The CSU1 dataset contains 408 images with a resolution of 4000 x 3000 pixels, covering an area of approximately 0.2 $km^2$, including the sports stadium and gymnasium areas of the new campus at Central South University. }
  \item{{\bf CSU2.} The CSU2 dataset comprises 713 images with a resolution of 4000 x 3000 pixels, covering approximately 1/3 of the southwest area of the new campus at Central South University. The total area covered is approximately 0.2 $km^2$. }
  \item{{\bf HNU.} The HNU dataset consists of 391 images with a resolution of 4000 x 3000 pixels, covering an area of approximately 0.1 $km^2$ around the Tianma Student Dormitory at Hunan University. }
  \item{{\bf CSUS.} The CSUS dataset consists of 777 captured photos with a resolution of 4000 x 3000 pixels, covering an on-site area of approximately 0.7 $km^2$ in the Southern Campus of Central South University. }
  \item{{\bf CSUHU.} The CSUHU dataset comprises 1706 photos covering an approximate area of 1 $km^2$. The main shooting locations include parts of the New Campus of Central South University and portions of the Houhu International Art Park buildings. The images in the dataset have a resolution of 4000 x 3000 pixels.}
\end{itemize}

\subsection{Results}
\subsubsection{Comparison methods}
In our experiments, the proposed method BirdNeRF is compared to four large-scale reconstruction solutions that can model with a single GPU:
\begin{itemize}
  \item{Metashape\cite{ref27}: Agisoft Metashape is a commercial software designed to process digital images using photogrammetry methods. We conducted our control experiments using Agisoft Metashape Pro 1.6.5.}
  \item{Mega-NeRF\cite{ref5}: Mega-NeRF is a scene segmentation method, but it performs segmentation at the pixel level. }
  \item{Instant-NGP\cite{ref13}: Instant-NGP is currently the fastest neural radiance field training method, and it serves as a benchmark for our approach BirdNeRF.} 
\end{itemize}

\subsubsection{Evaluation metrics}
We quantify the performance of our method using two established quantitative metrics: peak signal-to-noise ratio (PSNR) and structural similarity (SSIM). These metrics accurately assess the quality and similarity between the rendered images and their corresponding ground truth images. Additionally, we assess the efficiency of different methods by comparing the training time required after aligning the cameras in a consistent training environment. This dual evaluation framework provides a comprehensive analysis, addressing our proposed method's quality and efficiency dimensions.

\begin{figure*}[htbp]
    \centering
    \includegraphics[width=1\linewidth]{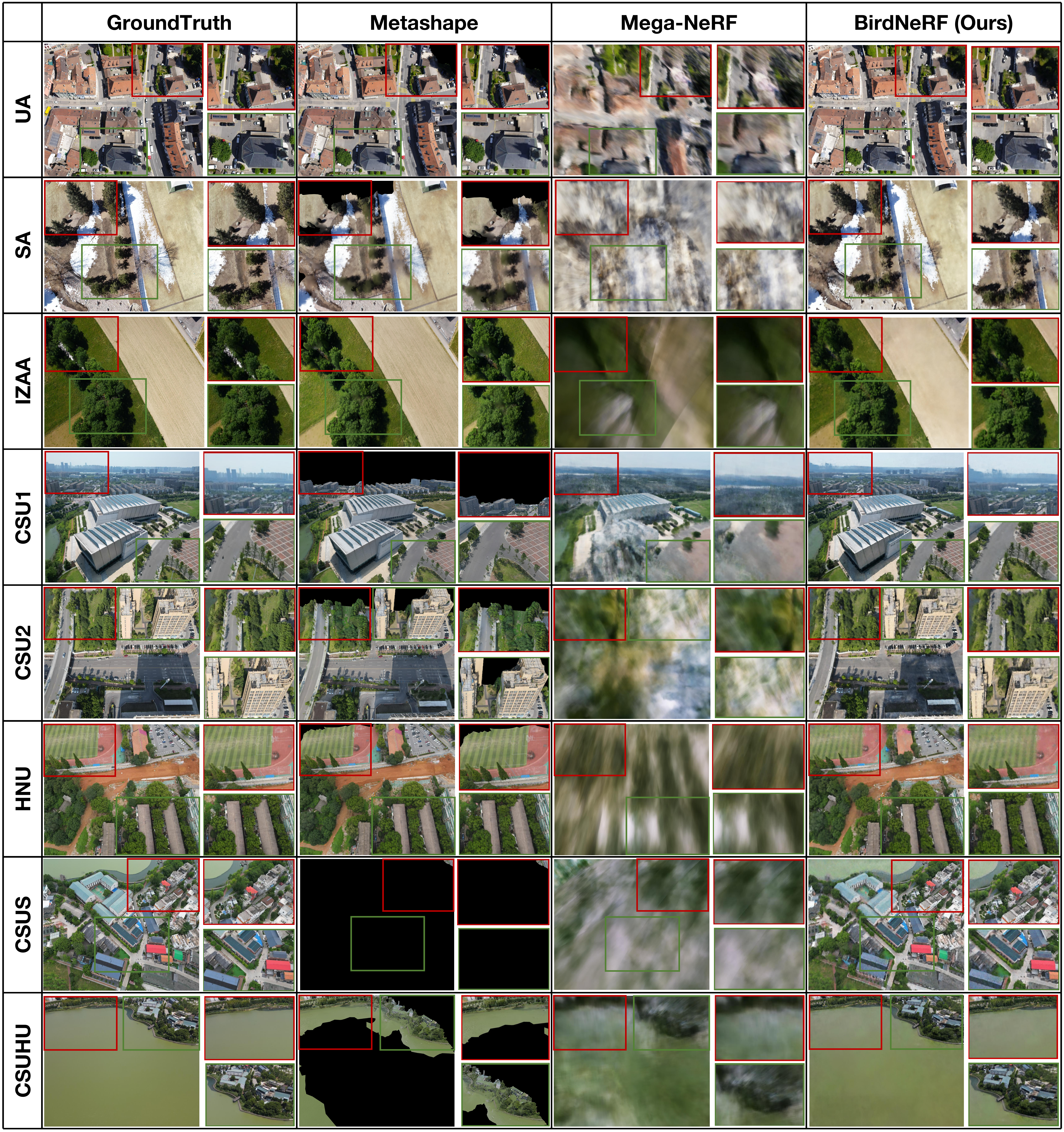} %
    \caption{Rendering outputs of Metashape, Mega-NeRF, and our method BirdNeRF.}
    \label{res-1}
\end{figure*}

\begin{figure*}[htbp]
    \centering
    \includegraphics[width=1\linewidth]{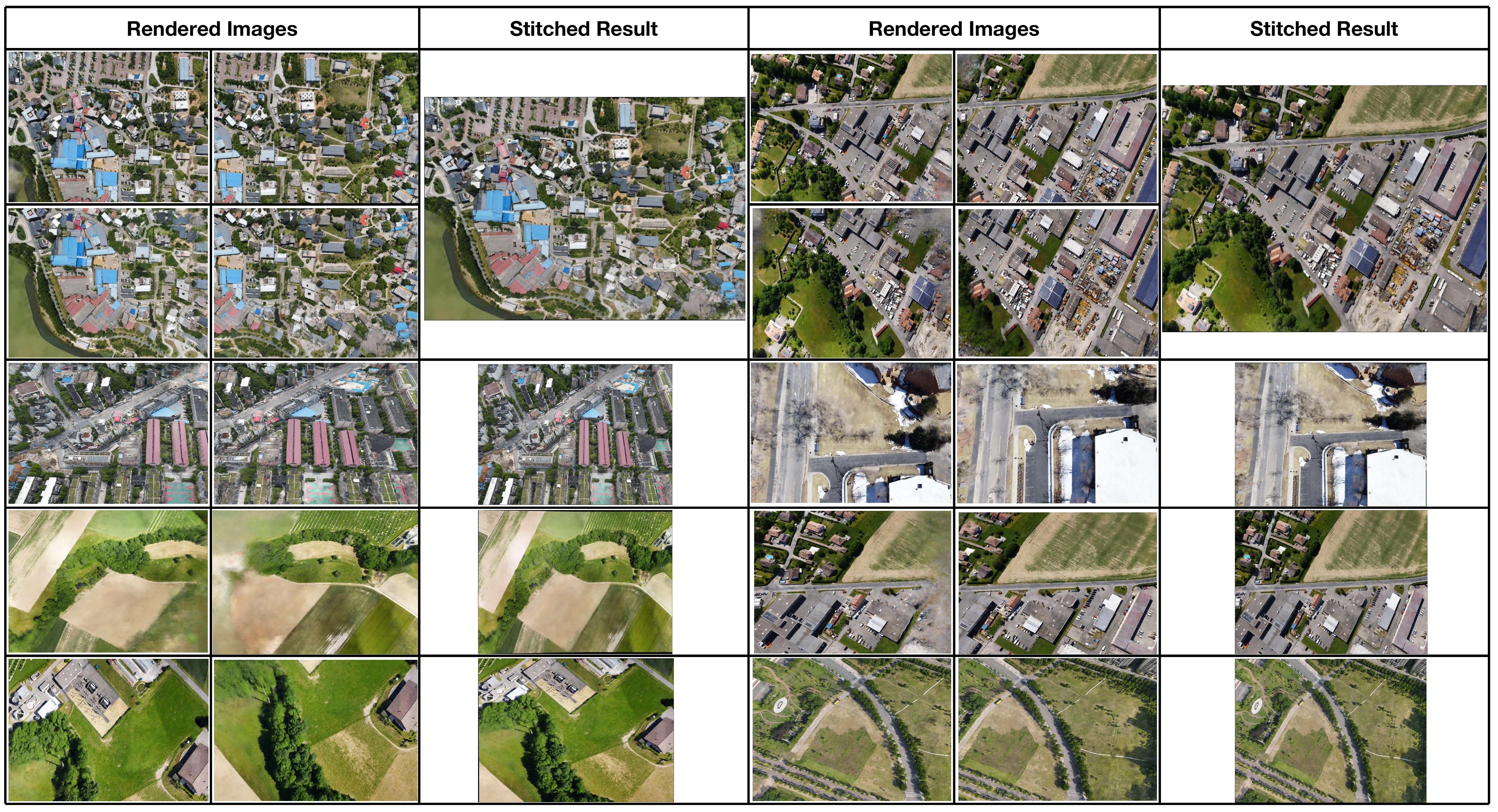}
    \caption{More results of randomly generated camera pose which may need a stitch.}
    \label{res-2}
\end{figure*}

\subsubsection{Qualitative results}
In Fig.~\ref{res-1}, we exhibit the post-modeling rendering outputs of the assessed methods. It is evident that Metashape's rendering output often contains undesirable holes and can lead to visually unrealistic objects. Mega-NeRF, on the other hand, exhibits subpar modeling results overall. In contrast, our proposed method demonstrates comparatively superior visual output results. Furthermore, we present additional results featuring randomly generated camera poses from our method, which may require stitching, as illustrated in Fig.~\ref{res-2}. This additional visualization provides insight into our approach's versatility and potential challenges in handling various camera poses.

\subsubsection{Quantitative analysis}
We conduct a quantitative analysis of the rendering results and training speed across different methods. Our findings reveal that our method excels in rapidly attaining high-quality results in large-scale 3D reconstruction, even under constraints of limited GPU memory resources.

{\bf PSNR and SSIM scores.} Table~\ref{tab:table0} depicts the PSNR and SSIM scores, offering a comparative analysis of the rendering results from all evaluated methods in this study against the ground truth images. Our method consistently achieves the highest PSNR scores across all datasets, securing SSIM superiority on nearly half of them, demonstrating greater stability in both metrics. Metascape displays a broader range of PSNR fluctuations. In contrast, our method maintains a consistently narrow range of fluctuations in scores, indicating its robustness and ability to deliver consistent and reliable results.

\begin{table*}[htbp]
    \caption{PSNR and SSIM scores of rendered results from Metashape\cite{ref27}, Mega-NeRF\cite{ref5},Instant-NGP\cite{ref13} and BirdNeRF. \label{tab:table0}}
    \centering
    \renewcommand\arraystretch{1.5}
     \setlength{\tabcolsep}{6mm}{ 
    \begin{tabular}{cccccccc}  
    \toprule  
    \multirow{2}*{\textbf{Dataset}}& \multirow{2}*{ Method} & \multicolumn{3}{c}{PSNR $\uparrow$ } & \multicolumn{3}{c}{SSIM $\uparrow$ } \\
    \cline{3-8} 
     ~ & ~  & Min & Max & Average  & Min & Max & Average   \\
    \cline{1-8}  
    
    \multirow{3}{*}{UA} & Metashape  & 13.619 & 24.713 & 19.891 & 0.890 & 0.963 & \bf{0.930} \\
    \multirow{3}{*}{} & Mega-NeRF  & 9.860 & 13.106 & 11.382 & 0.244 & 0.355 & 0.282\\
    \multirow{3}{*}{} & Instant-NGP  & 19.787 & 21.35& 20.154& 0.514 & 0.617 & 0.562  \\
    \multirow{3}{*}{} & \underline{BirdNeRF} & 19.603 & 20.768  & \bf{20.167} & 0.563 & 0.6560 & 0.595 \\
    \midrule
    \multirow{3}{*}{SA} & Metashape  & 5.135 & 34.819 & 18.517 & 0.686 & 0.995 & \bf{0.935 }\\
    \multirow{3}{*}{} & Mega-NeRF  & 7.105 & 12.514 & 9.808 & 0.146 & 0.469 & 0.296 \\
    \multirow{3}{*}{} & Instant-NGP  & $\backslash$ & $\backslash$ & $\backslash$ & $\backslash$ & $\backslash$ & $\backslash$  \\
    \multirow{3}{*}{} & \underline{BirdNeRF} & 16.508 & 23.120  & \bf{20.002} & 0.448 & 0.688 & 0.555 \\
    \midrule
    \multirow{3}{*}{IZAA} & Metashape  & 6.353 & 35.99 & 21.523 & 0.654 & 0.994 & \bf{0.887} \\
    \multirow{3}{*}{} & Mega-NeRF  & 8.757 & 19.116 & 12.845 & 0.124 & 0.548 & 0.276\\
    \multirow{3}{*}{} & Instant-NGP  & $\backslash$ & $\backslash$ & $\backslash$ & $\backslash$ & $\backslash$ & $\backslash$  \\
    \multirow{3}{*}{} & \underline{BirdNeRF} & 17.223 & 28.847 & \bf{21.829} & 0.399 & 0.775 & 0.544\\
    \midrule
    \multirow{3}{*}{CSU1} & Metashape  & 2.492 & 30.568 & 9.787 & 0.427 & 0.987 & 0.686  \\
    \multirow{3}{*}{} & Mega-NeRF  & 16.813 & 23.428 & 20.101 & 0.401 & 0.696 & 0.540  \\
    \multirow{3}{*}{} & Instant-NGP  & $\backslash$ &  $\backslash$ & $\backslash$  & $\backslash$ & $\backslash$ & $\backslash$   \\
    \multirow{3}{*}{} & \underline{BirdNeRF} & 18.226 & 26.255 & \bf{22.579} & 0.602 & 0.772 & \bf{0.689} \\
    \midrule
    \multirow{3}{*}{CSU2} & Metashape  & 5.974 & 31.373 & 19.167 & 0.725 &  0.985 & \bf{0.911}   \\
    \multirow{3}{*}{} & Mega-NeRF  & 8.887 & 15.577 & 12.308 &  0.125 & 0.395 & 0.257 \\
    \multirow{3}{*}{} & Instant-NGP  & $\backslash$ & $\backslash$ & $\backslash$ & $\backslash$ & $\backslash$ & $\backslash$   \\
    \multirow{3}{*}{} & \underline{BirdNeRF} & 11.013 & 25.290 & \bf{19.868} & 0.286 & 0.701 & 0.539 \\
    \midrule
    \multirow{3}{*}{HNU} & Metashape  & 5.806 & 28.251 & 21.127 & 0.723 & 0.988 & \bf{0.929}  \\
    \multirow{3}{*}{} & Mega-NeRF  & 10.236 & 13.852 & 11.937 & 0.188 & 0.319 & 0.252 \\
    \multirow{3}{*}{} & Instant-NGP  & $\backslash$ & $\backslash$ & $\backslash$ & $\backslash$ & $\backslash$ & $\backslash$  \\
    \multirow{3}{*}{} & \underline{BirdNeRF} & 17.664 & 22.954 & \bf{21.158} & 0.427 & 0.618 & 0.552 \\
    \midrule
    \multirow{3}{*}{CSUS} & Metashape  & 0.000 & 39.632 & 12.885 & 0.000 & 0.998 & 0.541 \\
    \multirow{3}{*}{} & Mega-NeRF  & 10.102 & 13.325 & 12.023 & 0.227 & 0.423 & 0.314 \\
    \multirow{3}{*}{} & Instant-NGP  & $\backslash$ & $\backslash$ & $\backslash$ & $\backslash$ & $\backslash$ & $\backslash$   \\
    \multirow{3}{*}{} & \underline{BirdNeRF} & 19.287 & 23.907 & \bf{21.580} & 0.448 & 0.659 & \bf{0.564} \\
    \midrule
    \multirow{3}{*}{CSUHU} & Metashape  & 0.957& 38.553 & 8.776 & 0.190 & 0.996 & 0.658 \\
    \multirow{3}{*}{} & Mega-NeRF  & 10.748 & 20.995 & 14.135 & 0.191 & 0.921 & 0.468 \\
    \multirow{3}{*}{} & Instant-NGP  & $\backslash$ & $\backslash$ & $\backslash$ & $\backslash$ & $\backslash$ & $\backslash$  \\
    \multirow{3}{*}{} & \underline{BirdNeRF} & 17.459 & 26.787 & \bf{21.505} & 0.442 & 0.889 & \bf{0.659} \\
    \bottomrule 
    \end{tabular}}
\end{table*}

{\bf Runtime comparison.} Instant-NGP demonstrates rapid convergence after $3 \times 10^3$ training iterations, whereas Mega-NeRF, an enhanced version of the initial NeRF, requires approximately $1 \times 10^5$ iterations to reach convergence. To ensure sufficient training for both Mega-NeRF and our method BirdNeRF, we set the training iterations for our method to $5 \times 10^3$ and for Mega-NeRF to $1 \times 10^5$ iterations. In Mega-NeRF's clustering mask partitioning stage, we set the grid dimension to 2x4, dividing each scene into 8 sub-scenes. Utilizing a single NVIDIA GeForce RTX 3090 GPU (24GB), the training environment allows for parallel training of two sub-scenes in Mega-NeRF. Additionally, the time Metashape consumes from sparse point cloud to mesh generation is considered the training time.

\begin{table}[!t]
    \caption{A comparison of training time for various methods. \label{tab:table3}}
    \centering
    \renewcommand\arraystretch{1.5}
    \begin{tabular}{cccc}
    \toprule  
    \multirow{2}*{Method} & \multicolumn{3}{c}{Training Time(h) $\downarrow$ }  \\
    \cline{2-4} 
     ~  & Metashape & Mega-NeRF & BirdNeRF   \\
    \cline{1-4} 
    UA & 0.041 & 17.109 & 0.014  \\
    SA & 0.253 & 17.413 &  0.029\\
    IZAA & 3.047 & 16.863  & 0.300  \\
    CSU1 & 0.445 & 17.238 & 0.076 \\
    CSU2 & 1.365 & 17.067 & 0.125  \\
    HNU & 0.842 & 18.178 & 0.059  \\
    CSUS & 2.304 & 30.748 & 0.154 \\
    CSUHU & 4.281 & 27.847 & 0.341  \\
    \bottomrule 
    \end{tabular}
    \end{table}

    \begin{figure}[!t]
    \centering
    \includegraphics[width=0.9\linewidth]{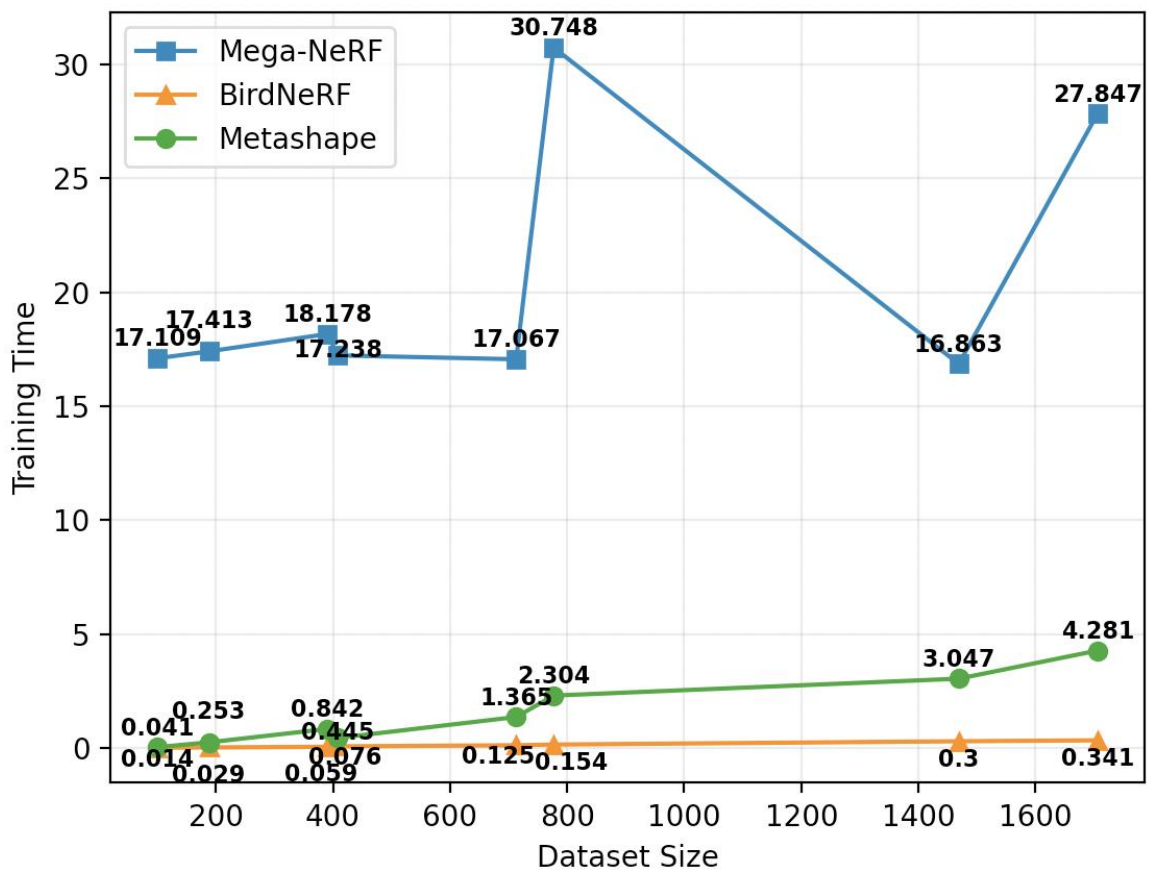}
    \caption{The figure above illustrates the trend of training time across all datasets. Our method shows a very gradual increase with the growth of data volume, and the larger the dataset, the more pronounced the advantage of our approach becomes.}
    \label{fig_time}
    \end{figure}

Table~\ref{tab:table3} provides a comprehensive comparison of training times for various methods, highlighting the significant speed advantage of our approach. Fig.~\ref{fig_time} visually represents this result, clearly illustrating the overall lengthy training times of Mega-NeRF, based on the naive NeRF implementation. Metashape's training time escalates quickly with the dataset size, whereas our method exhibits superior adaptability to large datasets with a slow increase in training time as the dataset grows.

\begin{table}[!t]
    \caption{The PSNR and SSIM scores for different training iterations of our method BirdNeRF. \label{tab:table4}}
    \centering
    \renewcommand\arraystretch{1.5}
    \begin{tabular}{ccccc}
    \toprule  
    ~ & Training iterations & Training time(h) $\downarrow$ & PSNR $\uparrow$ & SSIM $\uparrow$\\
    \midrule  
    \multirow{4}{*}{UA} & $5 \times 10^3$ & 0.014 & 20.167 & 0.595  \\
    \multirow{4}{*}{} & $1 \times 10^4$ & 0.029 & 20.432 & 0.610 \\
    \multirow{4}{*}{} & $3 \times 10^4$ & 0.093 & 20.830  & 0.636  \\
    \midrule
    \multirow{4}{*}{SA} & $5 \times 10^3$ & 0.029 & 20.002 & 0.555  \\
    \multirow{4}{*}{} & $1 \times 10^4$ & 0.063 & 20.236 & 0.568 \\
    \multirow{4}{*}{} & $3 \times 10^4$ & 11.667 & 20.725  & 0.590  \\
    \midrule  
    \multirow{4}{*}{IZAA} & $5 \times 10^3$ & 0.300 & 21.829 & 0.544  \\
    \multirow{4}{*}{} & $1 \times 10^4$ & 0.599 & 22.231 & 0.554 \\
    \multirow{4}{*}{} & $3 \times 10^4$ & 1.730 & 22.725  & 0.569  \\
    \midrule  
    \multirow{4}{*}{CSU1} & $5 \times 10^3$ & 0.076 & 22.579 & 0.690  \\
    \multirow{4}{*}{} & $1 \times 10^4$ & 0.149 & 23.306 & 0.713 \\
    \multirow{4}{*}{} & $3 \times 10^4$ & 0.461 &  23.766 & 0.738  \\
    \midrule  
    \multirow{4}{*}{CSU2} & $5 \times 10^3$ & 0.125 & 19.868 & 0.539  \\
    \multirow{4}{*}{} & $1 \times 10^4$ & 0.253 & 20.107 & 0.557 \\
    \multirow{4}{*}{} & $3 \times 10^4$ & 0.753 & 20.329  & 0.582  \\
    \midrule  
    \multirow{4}{*}{HNU} & $5 \times 10^3$ & 0.059 & 21.158 & 0.552  \\
    \multirow{4}{*}{} & $1 \times 10^4$ & 0.119 & 21.745 & 0.587 \\
    \multirow{4}{*}{} & $3 \times 10^4$ & 0.361 & 22.451  & 0.633  \\
    \midrule  
    \multirow{4}{*}{CSUS} & $5 \times 10^3$ & 0.154 & 21.580 & 0.564  \\
    \multirow{4}{*}{} & $1 \times 10^4$ & 0.310 & 21.987 & 0.582 \\
    \multirow{4}{*}{} & $3 \times 10^4$ & 0.890 & 22.832  & 0.627  \\
    \midrule  
    \multirow{4}{*}{CSUHU} & $5 \times 10^3$ & 0.341 & 21.504 & 0.659  \\
    \multirow{4}{*}{} & $1 \times 10^4$ & 0.671 & 21.763 & 0.670 \\
    \multirow{4}{*}{} & $3 \times 10^4$ & 1.927 & 22.132  & 0.687  \\
    \bottomrule 
    \end{tabular}
    \end{table}

The training time of our method remains independent of image resolution but is contingent on the number of training iterations. To showcase the effectiveness of our method, we compare training speed and rendering results on two datasets at different iteration counts. Table~\ref{tab:table4} presents the PSNR and SSIM scores for varying training iterations, revealing negligible differences in modeling performance. This flexibility allows us to adjust training iterations based on our time requirements.

\section{Conclusions}
This paper introduces BirdNeRF, a fast neural reconstruction method designed for processing a large number of aerial images. It stands out as the fastest large-scale reconstruction method to date, emphasizing efficient memory resource utilization and high rendering quality. The spatial decomposition strategy, grounded in camera distribution, empowers BirdNeRF to decompose and train scenes within specified memory constraints, underscoring its robust scalability. The projection-guided novel view Re-rendering strategy ensures accurate indexing of sub-scene bounding boxes and precise querying of related sub-models, ensuring high-quality rendering for diverse camera poses. Evaluation results indicate substantial advancements over classical photogrammetric software and the state-of-the-art large-scale NeRF solutions. BirdNeRF achieves a speed increase of more than ten times on a single GPU while maintaining commendable rendering quality. Positioned as a noteworthy contribution to the field, BirdNeRF offers practical solutions for critical challenges, significantly enhancing the speed, scalability, and visual realism of large-scale aerial scene reconstruction. This improvement is particularly vital for real-time applications such as disaster response.


\section*{Acknowledgments}
This work is supported by the National Key R\&D Program of China (No. 2022ZD0119003), Nature Science Foundation of China (No. 62102145),
and Jiangxi Provincial 03 Special Foundation and 5G Program (Grant No.
20224ABC03A05).


\newpage
\begin{IEEEbiography}[{\includegraphics[width=1in,height=1.25in,clip,keepaspectratio]{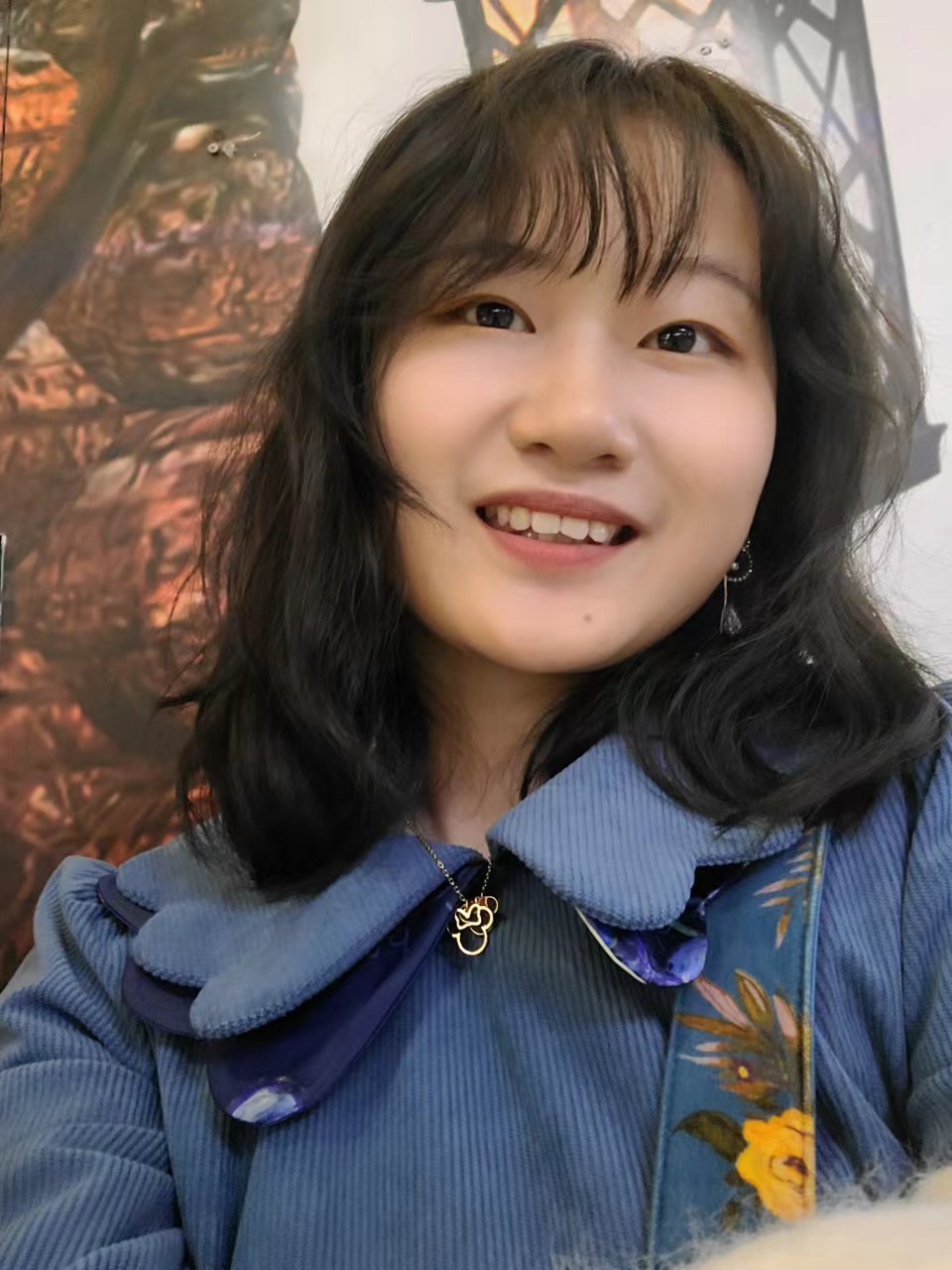}}]{Huiqing Zhang}
Huiqing Zhang is a master’s candidate at Hunan University, advised by Prof. Yizhen Lao, where she works in Computational Photography and 3D Computer Vision. She focuses on 3D Reconstruction and Neural Rendering. Before that, She got Bachelor’s Degree in Computer Science and Technology from Jiangxi Normal University. She previously led the Association of Computer Science at JXNU, and her undergraduate experience mainly focused on the Programming Contest.
\end{IEEEbiography}
\vspace{3pt}
\begin{IEEEbiography}
[{\includegraphics[width=1in,height=1.25in,clip,keepaspectratio]{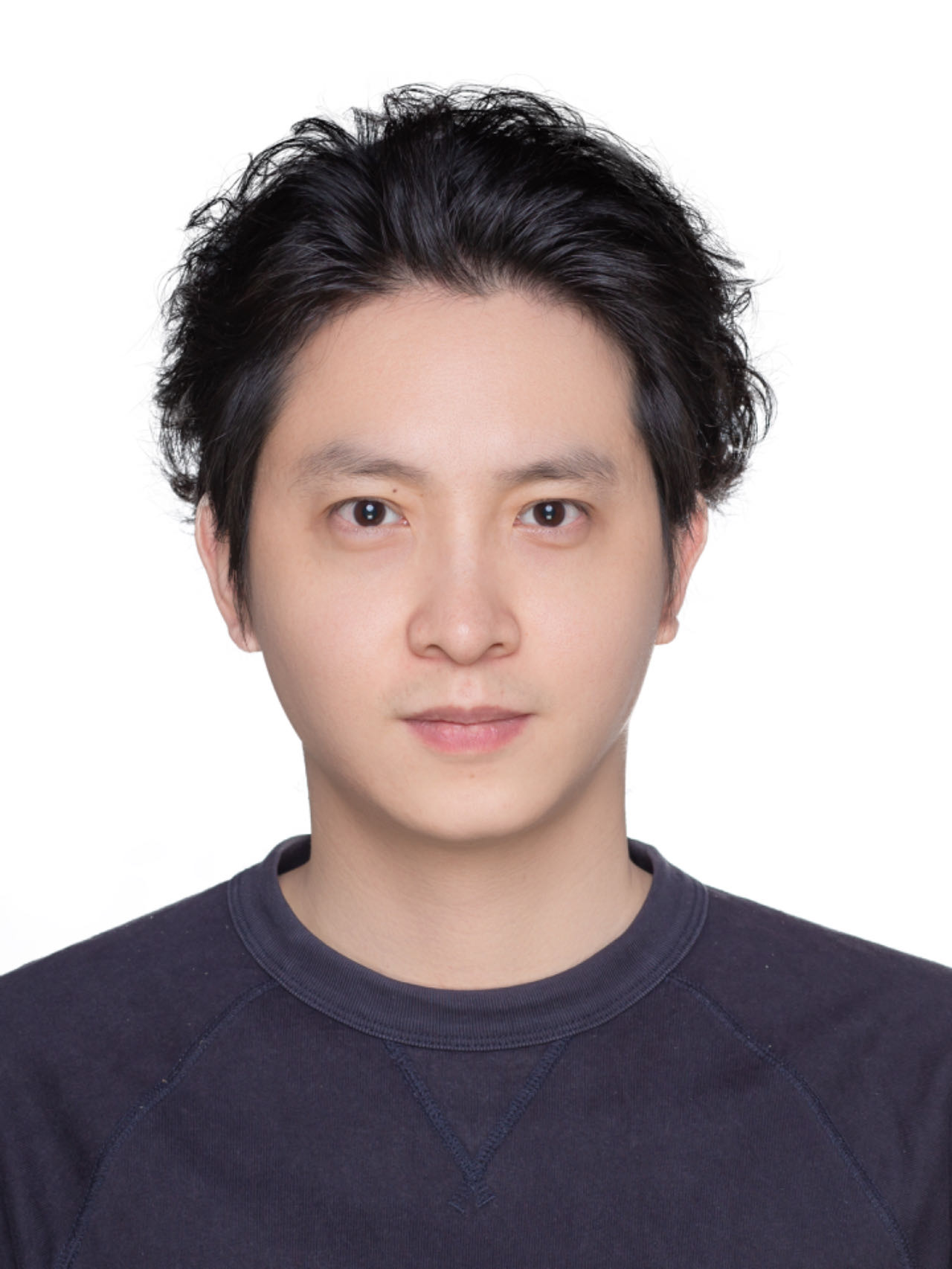}}]{Yifei Xue}
Yifei Xue received a Master of Science degree in surveying and mapping from the State Key Laboratory of Information Engineering in Surveying, Mapping, and Remote Sensing (LIESMARS) at Wuhan University, China, in 2014. Currently, he is a PhD candidate at Hunan University and holds the position as Head of the Office at Hunan Provincial Lushan Laboratory. The primary areas of his research include computer vision, remote sensing, computational photography, and machine learning.
\end{IEEEbiography}
\vfill
\newpage

\begin{IEEEbiography}
[{\includegraphics[width=1in,height=1.25in,clip,keepaspectratio]{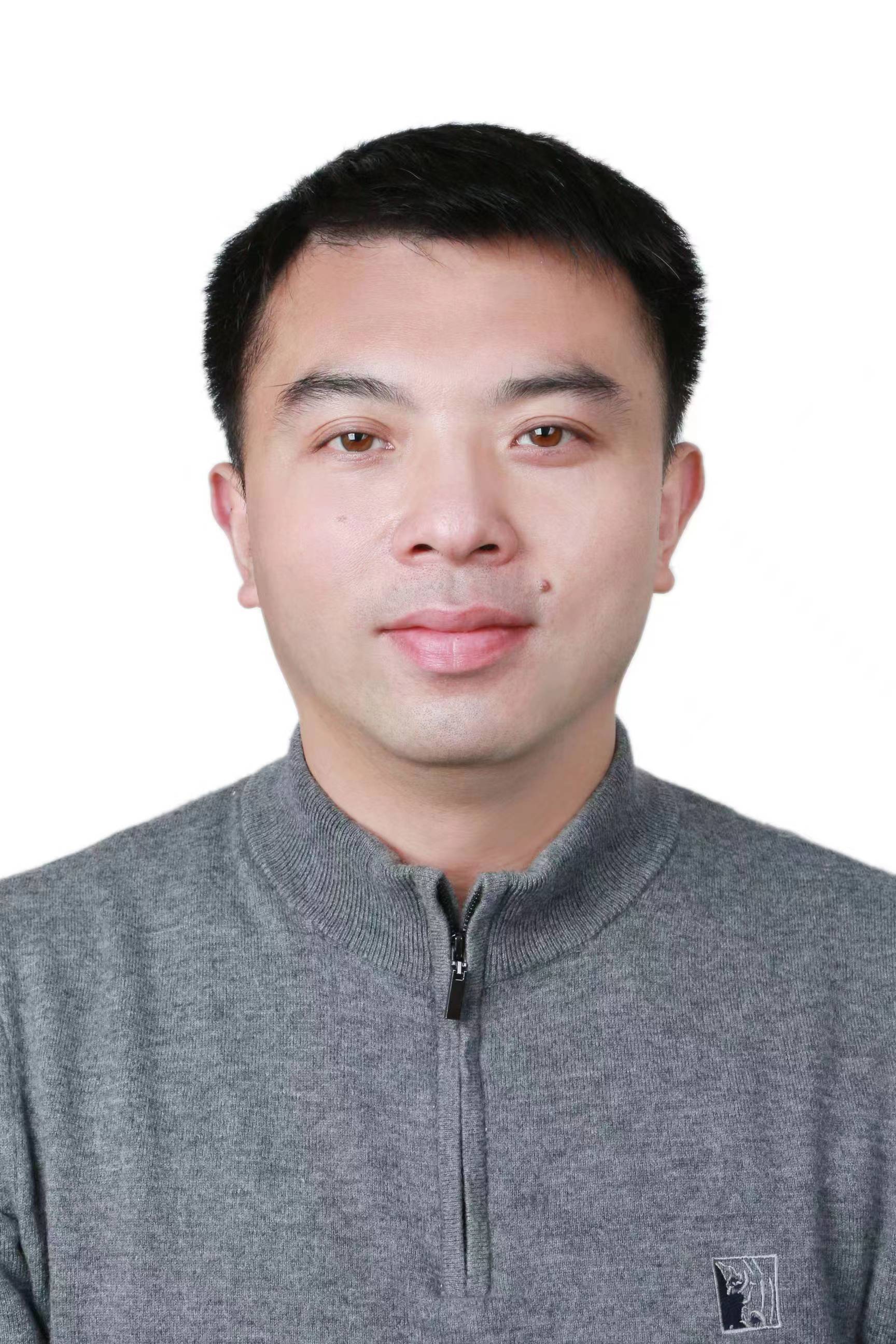}}]{Ming Liao}
Liao Ming, a professor-level senior engineer, Ph.D. graduated from Wuhan University, his main research includes natural resources informatization, and geographic information service platform construction. Currently, he is the deputy director of Jiangxi Provincial Natural Resources Cause Development Center and the executive deputy director of Jiangxi Province Engineering Research Center of Surveying, Mapping and Geographic Information.
\end{IEEEbiography}
\vspace{3pt}
\begin{IEEEbiography}
[{\includegraphics[width=1in,height=1.25in,clip,keepaspectratio]{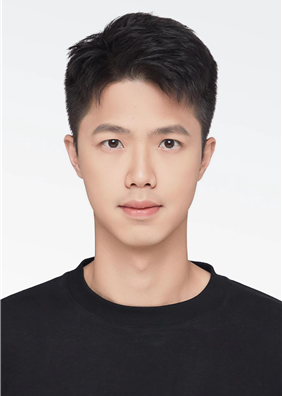}}]{Yizhen Lao}
Yizhen Lao (Member, IEEE) received the bachelor's degree in spatial-informatics and digitalized technology from Wuhan University, the master's degree in geo-information science and earth observation from the University of Twente, and the PhD degree from the University of Clermont-Auvergne. He is an associate professor with the College of Computer Science and Electronic Engineering, Hunan University. His research is focused on 3D computer vision and computational photography.
\end{IEEEbiography}


\vfill

\end{document}